\documentclass[runningheads]{llncs}

\usepackage{eccv}

\usepackage{eccvabbrv}

\usepackage{graphicx}
\usepackage{booktabs}

\usepackage[accsupp]{axessibility}  

\usepackage{booktabs}       
\usepackage{amsfonts}       
\usepackage{nicefrac}       
\usepackage{microtype}      
\usepackage{xcolor}         
\usepackage{colortbl}
\usepackage{amsmath}
\usepackage{tikz}
\usepackage{comment}
\usepackage{color}

\usepackage{enumitem}
\usepackage{graphicx}
\usepackage{amsmath}
\usepackage{booktabs}
\usepackage{mathtools}
\usepackage{url}
\usepackage{comment}
\usepackage{color}
\usepackage{graphicx,setspace}
\usepackage{multirow, array, booktabs,makecell,xspace,bm}
\usepackage{pifont}
\usepackage{amsfonts}
\usepackage{arydshln}
\usepackage{dsfont}
\usepackage{stackengine}
\usepackage{scalerel}
\usepackage{esdiff} 
\usepackage{bbm}

\newcommand{\cmark}{\ding{51}}%

\usepackage[most]{tcolorbox}

\usepackage{algorithm}
\usepackage{algpseudocode}

\usepackage{tikz}
\usepackage{comment}
\usepackage{color}
\usepackage{colortbl}

\usepackage{booktabs}       
\usepackage{amsfonts}       
\usepackage{nicefrac}       
\usepackage{microtype}      
\usepackage{xcolor}  

\usepackage{enumitem}
\usepackage{graphicx}
\usepackage{amsmath}
\usepackage{booktabs}
\usepackage{mathtools}
\usepackage{url}
\usepackage{comment}
\usepackage{color}
\usepackage{graphicx,setspace}
\usepackage{multirow, array, booktabs,makecell,xspace,bm}
\usepackage{pifont}
\usepackage{amsfonts}
\usepackage{arydshln}
\usepackage{dsfont}
\usepackage{stackengine}
\usepackage{scalerel}
\usepackage{esdiff} 
\usepackage{bbm}

\usepackage{multirow}

\usepackage{hyperref}

\usepackage{orcidlink}

\makeatletter
\newcommand{\blfootnote}[1]{%
  \begingroup
    \renewcommand\thefootnote{}%
    \renewcommand\@makefnmark{}
    \footnotetext{#1}%
  \endgroup
}
\makeatother

\begin{document}

\title{PersonaDrive: Controllable Trajectory Prediction with Multi-Dimensional Driving Personas} 

\titlerunning{PersonaDrive}

\author{Chan Lee\inst{1}\orcidlink{0009-0004-7827-8579}, Kimin Yun\inst{2}\orcidlink{0000-0002-4493-9437}, Yuseok Bae\inst{2}\orcidlink{0000-0002-4979-2649}, Seong Tae Kim\inst{1}$^{\dagger}$\orcidlink{0000-0002-2132-6021}, \\ and Jung Uk Kim\inst{1}$^{\dagger}$\orcidlink{0000-0003-4533-4875}}

\authorrunning{C.~Lee et al.}

\institute{Kyung Hee University, Yong-in, South Korea \\
\email{\{cksdlakstp12, st.kim, ju.kim\}@khu.ac.kr} \and
ETRI, Daejeon, South Korea\\
\email{\{kimin.yun, baeys\}@etri.re.kr}}

\maketitle

\blfootnote{$^{\dagger}$ Corresponding authors}

\begin{abstract}
    Although recent trajectory prediction and end-to-end autonomous driving methods improve robustness in urban environments, they still lack meaningful controllability. Existing benchmarks either provide no persona-conditioned annotations or support only a single urgency spectrum (\textit{i.e.,} emergency, normal, relaxed), which cannot distinguish personas that share the same urgency level but require different driving dynamics. To address this, we propose (\textit{i}) the Persona-Conditioned Trajectory (PCT) dataset, which decomposes driving personas along two axes—Temporal Urgency and Ride Comfort—and combines three levels of each to form a grid of nine personas, each paired with natural-language descriptions and trajectories, and (\textit{ii}) PersonaDrive, a framework that can learn driving personas from language and can generate persona-specific trajectories. PersonaDrive incorporates Persona-Conditioned Anchor Transform (PCAT), which hierarchically reshapes anchors along both axes, and Persona-Conditioned Multi-Modal Fusion (PCMF) for BEV-level persona fusion. Training is supervised by a Hierarchical Guide Loss enforcing axis-aligned physical orderings and an Axis-Decomposed Diversity Loss preventing diagonal mode collapse. \color{black}Experimental results show that PersonaDrive consistently improves over the compared baselines across multi-dimensional scenarios\color{black}. The code and PCT dataset are available at \url{https://github.com/VisualAIKHU/PersonaDrive}.

    \keywords{Autonomous Driving \and Dataset and Benchmark \and Multi-modal Learning}
\end{abstract}

\section{Introduction}
\label{sec:introduction}

As techniques that learn driving policies directly from raw sensor inputs--such as object detection \cite{chen2022polar, huang2021bevdet, li2024bevformer, wang2022detr3d, gu2021densetnt, shi2024mtr++, jung2026monosaod, lee2025multispectral, oh2024monowad}, object tracking \cite{zeng2022motr, zhang2021fairmot, zhang2022bytetrack}, and online mapping \cite{liao2022maptr, liao2025maptrv2, liu2023vectormapnet}--continue to advance, the importance of autonomous driving has become increasingly prominent. Among various autonomous driving tasks, trajectory prediction has gained significant attention as a core component of future planning \cite{phan2020covernet, ngiam2021scene, jiang2023motiondiffuser}. This task leverages spatiotemporal information from the current scene, such as camera and LiDAR observations, to forecast the future positions of the ego vehicle over upcoming time steps.

This importance of trajectory prediction is further highlighted by recent autonomous driving studies that have advanced the field through unified BEV representations \cite{hu2023planning}, multi-sensor fusion \cite{chitta2022transfuser}, anchor-based distillation \cite{li2024hydra, Weng2024paradrive}, and diffusion- or distribution-based prediction \cite{liao2025diffusiondrive, chen2024vadv2}.

\begin{figure}[t]
    \begin{minipage}[b]{\linewidth}
	\centering
        \centerline{\includegraphics[width=0.8\linewidth]{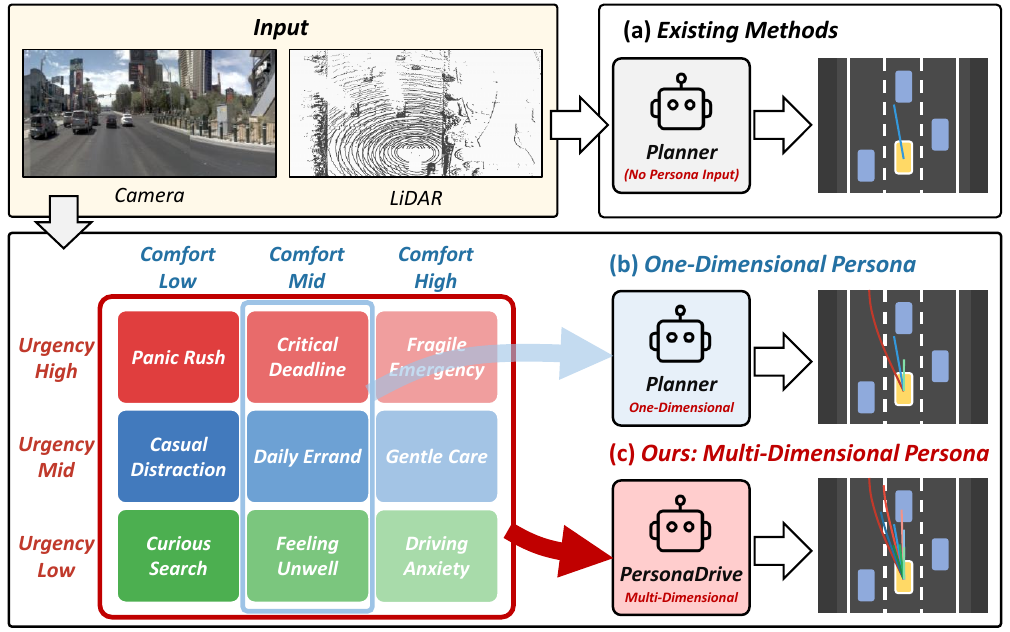}}
        \end{minipage}
	\caption {Conceptual comparison of (a) existing trajectory predictors, which output a single persona-agnostic plan, (b) one-dimensional persona methods conditioned on a single urgency axis, and (c) our PersonaDrive, which conditions on natural-language persona descriptions spanning Temporal Urgency and Ride Comfort, yielding nine distinct driving personas.}
    \label{fig:conceptual_comp}
    \vspace{-0.5cm}
\end{figure}

Despite these recent advancements, as shown in Figure \ref{fig:conceptual_comp}(a), many current trajectory prediction models still fall short in reflecting diverse human driving personas. As they are trained on datasets collected under fixed, normal driving conditions, these models often learn a single conservative trajectory pattern instead of adapting to diverse user personas. In reality, driving behavior varies widely depending on the driving situation, shaped by multiple concurrent factors including trip urgency, passenger comfort requirements, cargo fragility, road familiarity, and emotional state that interact to produce a unique driving style in each situation \cite{jurecki2021methodology, lajunen1997speed, peng2024drivers, kou2025padriver, hao2025styledrive, hasenjager2017personalization,liao2024bat}. Without controllable persona inputs, existing models cannot capture this behavioral diversity.

As illustrated in Figure \ref{fig:conceptual_comp}(b), prior work on persona-conditioned planning has attempted to address this gap by introducing a single spectrum of coarse categories such as emergency, normal, and relaxed \cite{chang2025langtraj,felemban2026imotion,wan2024tlcontrol,xia2024language,xu2024drivegpt4,hasenjager2017personalization,li2025learning,mao2023gpt}. However, such a one-dimensional formulation maps all personas within the same urgency level to identical outputs: for instance, a paramedic rushing a critical patient to a hospital and a firefighter racing to a blaze are both urgent, yet the former demands smooth, jerk-free motion to protect the patient while the latter tolerates aggressive maneuvering. Mapping both to a single ``emergency'' label erases this distinction and limits the advantage of using natural language over a simple one-hot encoding. 

To capture these orthogonal aspects of driving behavior, as shown in Figure \ref{fig:conceptual_comp}(c), we decompose persona into two independent and representative axes-Temporal Urgency and Ride Comfort-and discretize each into three levels to form nine distinct persona categories that span the full behavioral space.

In this paper, we introduce the Persona-Conditioned Trajectory (PCT) Dataset, which encodes driving persona along two behavioral axes—Temporal Urgency and Ride Comfort—yielding nine distinct persona categories (Figure \ref{fig:conceptual_comp}(c)). Each category is paired with natural-language persona descriptions and corresponding ground-truth trajectories for the same scenes. We further propose PersonaDrive, a framework that learns persona-driven behavior from language and generates persona-specific future trajectories. \color{black}By crossing three urgency and three comfort levels, the nine categories span the full spectrum of driving personas, from someone rushing to an urgent appointment regardless of discomfort, to someone delivering fragile instruments who needs the smoothest possible ride.\color{black}

Notably, scenarios such as the urgent-but-gentle need of a persona transporting delicate equipment and the urgent-but-rough tolerance of a persona racing to a critical meeting share the same urgency level yet require different trajectories, making text-based conditioning essential beyond simple categorical labels \cite{moon2024visiontrap}. We address two key challenges: (\textit{i}) how to control predictions by integrating the text-based persona description with the current scene across nine diverse categories, and (\textit{ii}) how to generate diverse, persona-conditioned trajectories while maintaining clear separation along both the urgency and comfort axes.

To address the challenge (\textit{i}), we introduce Persona-Conditioned Multi-Modal Fusion (PCMF), which aligns persona cues with visual and spatial representations through query-conditioned compression and gated fusion, enabling persona-aware trajectory prediction. We also propose Persona-Conditioned Anchor Transform (PCAT), which modulates the base anchors that serve as initial trajectory prototypes according to the target persona. Together, these components enable trajectories to respond to persona changes across both axes in a stable, predictable way, improving controllability.

\color{black}For challenge (\textit{ii}), we devise a diversity loss whose core idea is to preserve the persona separation observed in ground-truth behaviors along both axes, penalizing predicted trajectories that converge despite different personas. As a result, our framework enables direct control of diverse, persona-conditioned trajectory predictions via natural-language descriptions, differentiating multi-dimensional persona scenarios that one-dimensional labels cannot.\color{black}

The major contributions can be summarized as follows:
\begin{itemize}
\item We introduce a two-axis persona formulation that decomposes driving behavior along Temporal Urgency and Ride Comfort, yielding nine distinct persona categories, and construct the PCT dataset that provides natural-language persona descriptions paired with corresponding ground-truth trajectories for all nine categories.
\item We propose PersonaDrive, a framework that learns driving persona from text and injects it into planning through PCAT and PCMF.
\item \color{black}Experiments on the NAVSIM closed-loop benchmark show that PersonaDrive improves over the compared baselines across all nine persona categories for persona-aware trajectory prediction.\color{black} 
\end{itemize}

\section{Persona-Conditioned Trajectory Dataset}
\label{sec:dataset}

\subsection{Multi-Dimensional Behavioral Decomposition}
\label{sec:decomposition}
People vary driving behavior according to trip purpose and current state \cite{jurecki2021methodology, lajunen1997speed, peng2024drivers}, yet existing trajectory prediction datasets such as OpenScene do not provide controllable inputs \cite{openscene2023,wei2025pdb,caesar2020nuscenes,caesar2021nuplan,ettinger2021large,wilson2023argoverse}, causing models to converge to average behavior rather than learning persona-conditioned distributions. To move beyond a single urgency axis (\textit{e.g.}, emergency / normal / relaxed), we define two orthogonal behavioral axes that jointly form a multi-dimensional persona space: \\

\noindent \textbf{(\textit{i}) Temporal Urgency (Urgency)}: the degree of time pressure perceived by the driver, which primarily governs speed and forward progress. High urgency yields faster travel and decisive maneuvers; low urgency allows leisurely pacing. \\

\noindent \textbf{(\textit{ii}) Ride Comfort (Comfort)}: the priority placed on ride smoothness and passenger well-being, which primarily governs acceleration profiles and lateral dynamics. High comfort demands gentle acceleration, smooth cornering, and minimal jerk; low comfort tolerates aggressive dynamics. \\

These two axes are orthogonal: urgency dictates how fast the vehicle should progress, while comfort dictates how smoothly it should do so. By discretizing each axis into three levels (High / Mid / Low), we obtain a 3$\times$3 grid of nine personas, as illustrated in Figure \ref{fig:conceptual_comp}. This grid structure enables rigorous validation of axis independence: by fixing one axis and varying the other, we can verify that each dimension contributes distinct and measurable behavioral changes to the predicted trajectory (Section \ref{sec:decomposition}).

\begin{figure*}[t] 
  \centering
  \includegraphics[width=\linewidth]{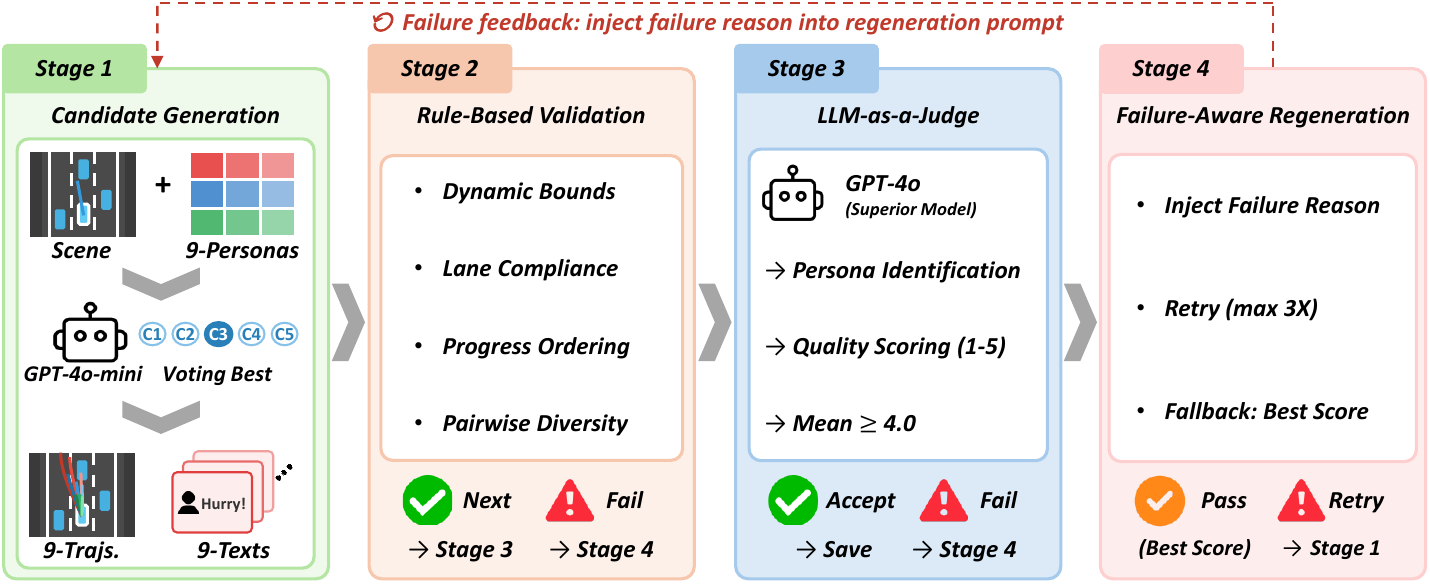}
    \caption{Four-stage generation and validation pipeline for the PCT dataset. Stage 1 generates five candidate sets of nine trajectory parameters via GPT-4o-mini and selects the best through voting. Stages 2 and 3 run in parallel: rule-based validation checks dynamic bounds, lane compliance, progress ordering, and pairwise diversity, while GPT-4o evaluates trajectory quality and verifies persona consistency. Samples failing either stage are regenerated in Stage 4, where failure context is injected into the prompt.}
  \label{fig:generation_pipeline}
  \vspace{-0.5cm}
\end{figure*}

\subsection{Dataset Construction} 
The PCT dataset represents each persona using natural-language descriptions that take the form of a passenger request to an autonomous taxi, naturally encoding both urgency and comfort through factors such as age, trip purpose, stress level, and emotional state. For example, a panicked passenger desperately rushing to save an overdosed friend, unconcerned about ride quality, maps to (Urgency High, Comfort Low), while a parent urgently transporting a severely burned child who cannot tolerate any bumps maps to (Urgency High, Comfort High). By varying these factors, the text spans the full $3\times3$ persona grid without requiring explicit axis labels.

\begin{figure*}[t] 
  \centering
  \includegraphics[width=0.7\linewidth]{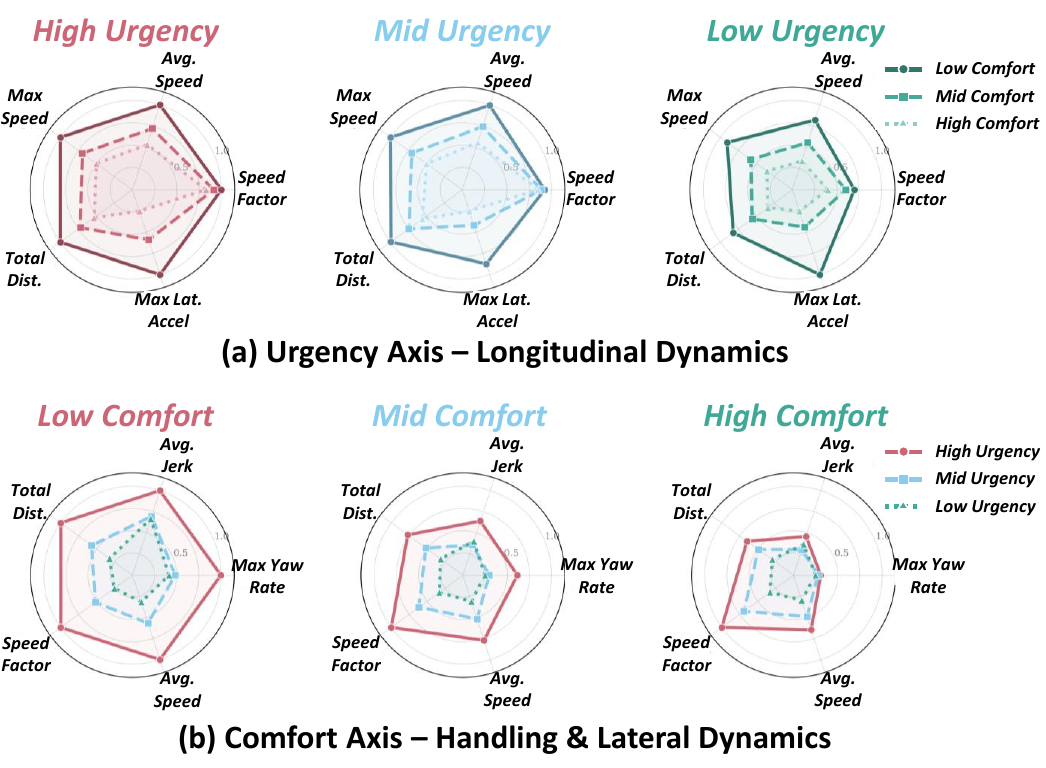}
    \caption{Multi-dimensional behavioral profiles of the nine personas. (a) With urgency fixed, longitudinal metrics (speed, distance) separate by urgency level while comfort lines overlap. (b) With comfort fixed, handling metrics (jerk, yaw rate) separate by comfort level while urgency lines overlap. This confirms that the two axes control orthogonal aspects of driving behavior.}
  \label{fig:dataset_analysis_radar}
\end{figure*}

\subsection{Generation and Validation Pipeline} 
We construct the dataset through a four-stage pipeline illustrated in Figure \ref{fig:generation_pipeline}. For each scene, structured scene information (lane geometry, drivable area, surrounding objects, traffic signals, ego state) is extracted into a JSON representation, and a system prompt specifies the nine persona definitions and trajectory generation parameters. The actual waypoint coordinates are computed from these parameters by applying lane-keeping constraints and drivable-area clamping. \\

\noindent \textbf{Stage 1: Candidate Generation and Voting.}
For each scene, we generate five candidate sets of nine trajectory parameter sets using GPT-4o-mini and select the best through a voting mechanism that considers plausibility and diversity. The corresponding passenger request texts are then generated via a separate API call, grounded in the actual trajectory characteristics.\\

\noindent \textbf{Stage 2: Rule-Based Validation.}
The selected candidate undergoes deterministic checks: dynamic bounds on speed, acceleration, and jerk; adherence to the $3\times3$ progress pattern; TTC $\geq 1.0$\,s; and minimum pairwise diversity (ADE $\geq 0.3$\,m). For texts, we verify sufficient diversity via Jaccard similarity.\\

\noindent \textbf{Stage 3: LLM-as-a-Judge.}
In parallel with Stage 2, GPT-4o \cite{achiam2023gpt}—a more capable model than the GPT-4o-mini generator—evaluates each candidate following the LLM-as-a-Judge 
paradigm \cite{zheng2023judging}. A trajectory judge scores realism, safety, and persona consistency, while a text judge classifies each description into its correct persona category. A sample passes only when the combined score exceeds a threshold.\\

\noindent \textbf{Stage 4: Failure-Aware Regeneration.}
Scenes that fail validation are regenerated with failure reasons injected into the prompt. Rule-based failures allow up to three rounds and quality-based failures up to two. If a scene still does not pass, the result with the highest combined score is selected. We applied this pipeline to the entire OpenScene dataset \cite{openscene2023}, generating nine persona-conditioned trajectory-text pairs for every scene in both the train and test splits. \color{black}To verify that the resulting persona labels are not artifacts of the GPT-family judge, we re-judge a held-out sample with two independent model families (Claude, Gemini), which show substantial-to-near-perfect agreement ($\kappa=0.978/0.804$); details are in the supplementary material.\color{black}

\subsection{Dataset Analysis} 
Figure \ref{fig:dataset_analysis_radar}(a) and Figure \ref{fig:dataset_analysis_radar}(b) visualize the trajectory statistics of the generated dataset using radar charts, organized along each axis to verify that the two dimensions produce the expected behavioral signatures.\\

\noindent \textbf{Urgency axis (Figure \ref{fig:dataset_analysis_radar}(a))}
Each chart fixes one urgency level and overlays the three comfort levels. The longitudinal metrics (Avg. Speed, Max Speed, Speed Factor, Total Dist.) decrease consistently from High Urgency to Low Urgency, confirming that urgency primarily governs forward progress. In contrast, Max Lat. Accel shows clear separation across comfort levels: Low Comfort yields the largest lateral acceleration, while High Comfort compresses it, reflecting smoother cornering under higher comfort priority.\\

\noindent \textbf{Comfort axis (Figure \ref{fig:dataset_analysis_radar}(b)).}
Each chart fixes one comfort level and overlays the three urgency levels. The handling metrics (Avg. Jerk, Max Yaw Rate, Max Lat. Accel) decrease consistently from Low Comfort to High Comfort, confirming that comfort governs ride smoothness. In contrast, Avg Speed, Speed Factor, and Total Dist. show clear separation across urgency levels: High Urgency consistently occupies the outermost region while Low Urgency stays near the center.

These patterns jointly confirm axis independence: urgency controls longitudinal dynamics (speed, distance) while comfort controls lateral dynamics (jerk, yaw rate, lateral acceleration). Within each chart, the lines for the non-varying axis largely overlap, indicating that each axis varies independently without interfering with the other. A Wilcoxon signed-rank test \cite{wilcoxon1945individual} on the per-scene mean speed differences yields $p < 0.001$ for all urgency-adjacent pairs (within each comfort level), and a corresponding test on mean absolute jerk yields $p < 0.001$ for all comfort-adjacent pairs (within each urgency level), confirming statistical significance along both axes.

\begin{table*}[t]
    \renewcommand{\tabcolsep}{4mm}
    \centering
    \caption{Persona distinguishability of the PCT dataset. Each cell reports the accuracy (\%) for identifying the correct persona from shuffled trajectory--text pairs.}
    \resizebox{0.85\linewidth}{!}{
      \begin{tabular}{c c c c}
        \Xhline{3\arrayrulewidth}
        \rule{0pt}{9pt} & \bf Comf.\ Low & \bf Comf.\ Mid & \bf Comf.\ High \\ \hline
        \rule{0pt}{9pt}
        \bf Urg.\ High & 89.9 & 89.9 & 92.2 \\
        \bf Urg.\ Mid  & 98.4 & 82.2 & 82.9 \\
        \bf Urg.\ Low  & 94.6 & 73.6 & 67.4 \\
        \Xhline{3\arrayrulewidth}
      \end{tabular}
    }
\label{table:human_eval_distinguish}
\vspace{-0.4cm}
\end{table*}

\subsection{Human Evaluation} 
To assess the human plausibility of the generated annotations, we conduct a user study with 13 participants. Each participant was presented with shuffled trajectory--text pairs from the same scene and asked to identify which of the nine personas each pair belongs to. As shown in Table \ref{table:human_eval_distinguish}, the overall accuracy is 85.7\%, highest for scenarios at the extremes of one or both axes and relatively lower for cells near the center of the grid, which are closer to default driving behavior. Most importantly, accuracy remains well above chance even for cells sharing one axis value, demonstrating that humans can perceive the comfort dimension even when urgency is held constant and confirming that both axes carry distinguishable information. A separate quality and realism study is also conducted, where participants consistently rate both trajectory realism and text clarity favorably; detailed results are provided in the supplementary material.

\section{Proposed Method: PersonaDrive}
\label{sec:method}

\begin{figure*}[t] 
  \centering
  \includegraphics[width=\linewidth]{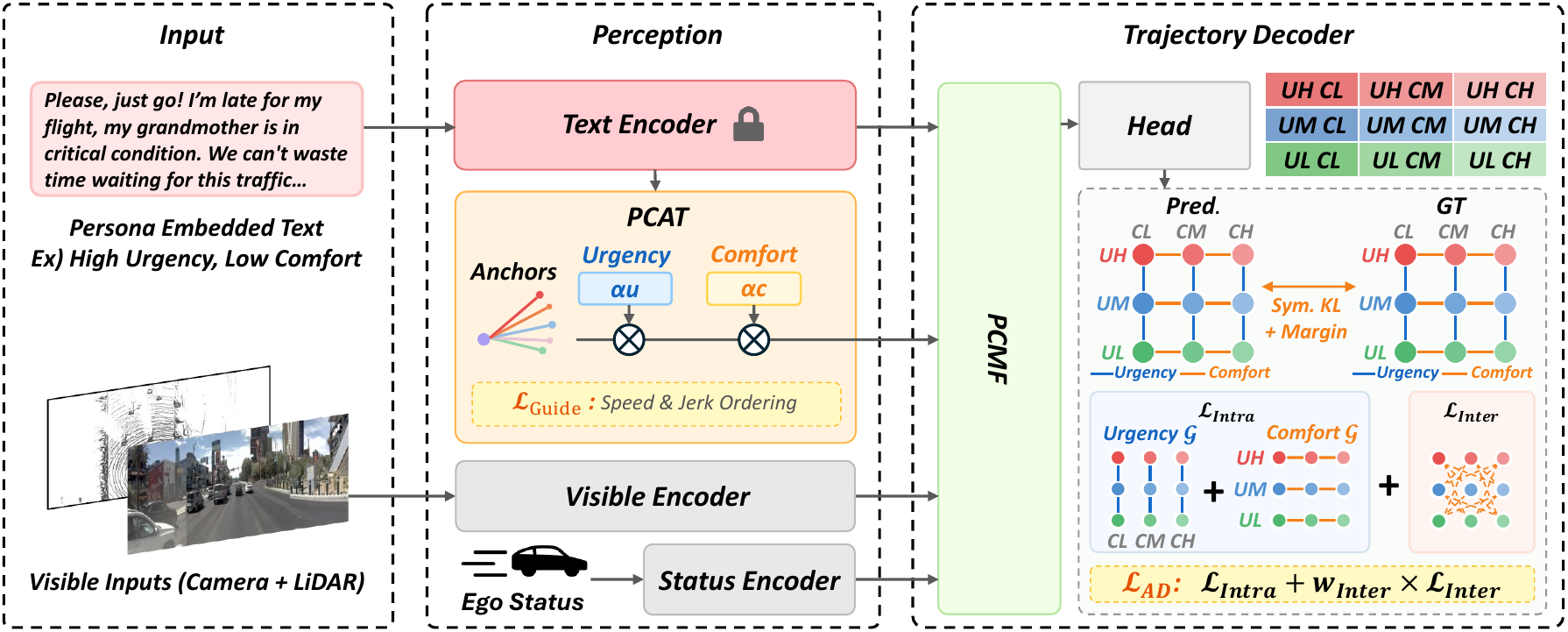}
    \caption{Overall architecture of the proposed PersonaDrive framework. PCAT modulates the anchor bank along the urgency and comfort axes using scalars derived from the text query. PCMF fuses BEV, ego, persona, and transformed anchor queries to produce persona-aware trajectories. Training is supervised by a Hierarchical Guide Loss and an Axis-Decomposed Diversity Loss. $\otimes$ denotes element-wise multiplication.}
  \label{fig:Overall_architecture}
  \vspace{-0.4cm}
\end{figure*}

Figure \ref{fig:Overall_architecture} shows the overall architecture. The inputs comprise cameras $I_{cam}$, LiDAR $I_{lidar}$, and a text description $T$ encoding the user driving persona. A visual encoder produces BEV queries $Q_{bev}$, and a frozen text encoder (\textit{e.g.,} MiniLM \cite{wang2020minilm}) produces $Q_{txt}$. PCAT receives $Q_{txt}$ and hierarchically modulates the trajectory anchor set $\mathcal{P}$ to generate the persona-adapted anchor set $\widetilde{\mathcal{P}}$. Together with the ego query $Q_{ego}$, $Q_{bev}$, $Q_{txt}$, and $\widetilde{\mathcal{P}}$ are fed to the PCMF module, and the head network estimates the persona-conditioned trajectory.

\subsection{Persona-Conditioned Anchor Transform}
To estimate a trajectory, a trajectory anchor set $\mathcal{P} = \{p_1, \dots, p_K\}$ serves as initial priors for prediction. Since the target persona can vary widely, we propose PCAT, which applies a hierarchical two-stage anchor modulation.

We decompose the text query $Q_{txt}$ into two independent branches. The urgency branch produces a scalar $\alpha_u$ that controls the global trajectory scale (\textit{i.e.,} overall travel distance), and the comfort branch produces a per-timestep modulation vector $\alpha_c \in \mathbb{R}^{T}$ that adjusts the local trajectory shape at each waypoint. Each branch consists of a two-layer MLP with GELU activation, followed by a task-specific projection:
\begin{equation}
    \alpha_u = f_u(Q_{txt}) \in \mathbb{R}, \quad \alpha_c = \sigma\!\bigl(f_c(Q_{txt})\bigr) \times \Delta_c + \gamma_c \in \mathbb{R}^{T},
\end{equation}
where $f_u$ and $f_c$ denote the urgency and comfort projection networks, and $\sigma$ is the sigmoid function. The comfort modulation $\alpha_c$ is bounded to $[\gamma_c,\; \gamma_c + \Delta_c]$ via the sigmoid-affine transformation. We set $\gamma_c = 0.8$ and $\Delta_c = 0.4$, constraining $\alpha_c$ to $[0.8, 1.2]$ for stable yet meaningful shape variation, while $\alpha_u$ is unconstrained for flexible global scaling.

The persona-adapted anchor set $\widetilde{\mathcal{P}} = \{\tilde{p}_1, \dots, \tilde{p}_K\}$ is then obtained by applying the two modulations sequentially. We first scale the entire anchor by $\alpha_u$ and then modulate each timestep independently by $\alpha_c$:
\begin{gather}
    \tilde{p}_i^{(u)} = p_i \times \alpha_u,  \quad
    \tilde{p}_i(t) = \tilde{p}_i^{(u)}(t) \times \alpha_c(t). 
\end{gather}
where $t$ indexes the timestep. For the neutral persona (medium urgency, medium comfort), we skip the transformation and use the base anchor $p_i$ as-is, which guarantees that the model behavior is identical to the baseline when no persona modification is applied.

\begin{figure}[t]
    \begin{minipage}[b]{\linewidth}
	\centering
        \centerline{\includegraphics[width=0.85\linewidth]{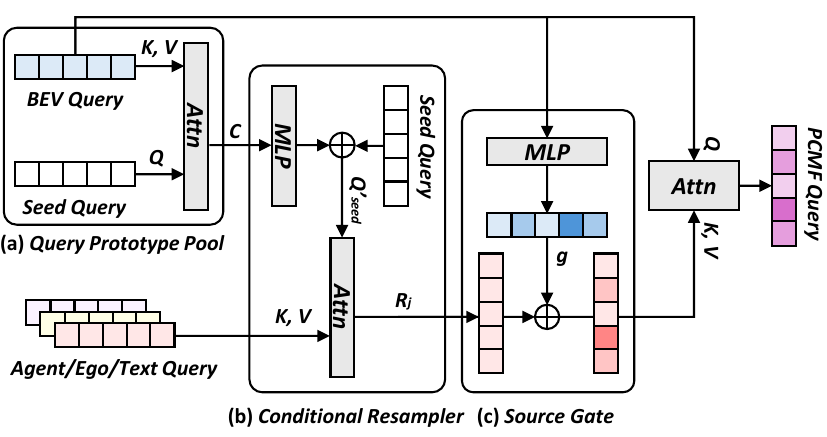}}
        \end{minipage}
	\caption {Illustration of the PCMF module. (a) Query Prototype Pool compresses $Q_{bev}$ into a conditional vector $C$. (b) Conditional Resampler modulates learnable slots with $\mathbf{C}$ and resamples each source. (c) Source Gate applies persona-dependent gating, and the PCMF fuser combines the resulting slots with BEV features.}
    \label{fig:PCMF}
    \vspace{-0.3cm}
\end{figure}

\subsection{Persona-Conditioned Multi-Modal Fusion}
After modulating the anchors, PCMF updates the input queries to reflect the user driving persona. We treat $Q_{bev}$ as the main query and combine it with $\widetilde{\mathcal{P}}$ to generate the agent query \color{black}$Q_{\text{agent}}$\color{black}. As shown in Figure \ref{fig:PCMF}, the module proceeds in three steps. First, the Query Prototype Pool compresses $Q_{bev}$ into a conditional vector via a set of $M{=}8$ learnable seed queries:
\begin{align}
    C = \mathrm{Attn}(Q_{\text{seed}}, Q_{bev}, Q_{bev}).
\end{align}
Next, the \textbf{Conditional Resampler} modulates learnable slots with a projected context vector $\tilde{C}$ and resamples each source into persona-aligned representations:
\begin{gather}
  Q_{slot}' = Q_{seed} + \tilde{C}, \quad
  R_j = \mathrm{Attn}(Q_{slot}', Q_j, Q_j), \; j \in \{\text{agent}, \text{ego}, \text{text}\},
\end{gather}
where $Q_{slot}'$ is obtained by adding $\tilde{C}$ to learnable slot seeds $Q_{seed}$, enabling the conditioned slots to attend selectively to tokens consistent with the target persona. A \textbf{Source Gate} then modulates source-wise contributions with a scene-conditioned gating vector $g = \mathrm{softmax}(\text{MLP}(Q_{bev}))$, yielding $\tilde{R}_j = g_j R_j$, where $g = [g_{\text{agent}}, g_{\text{ego}}, g_{\text{text}}]$ dynamically adjusts each source's contribution based on the scene complexity and target persona. Finally, the PCMF Fuser integrates the gated slots:
\begin{gather}
    Q_{PCMF} = \mathrm{Attn}\!\bigl(Q_{bev},\; [\tilde{R}_{\text{agent}}; \tilde{R}_{\text{ego}}; \tilde{R}_{\text{text}}],\; [\tilde{R}_{\text{agent}}; \tilde{R}_{\text{ego}}; \tilde{R}_{\text{text}}]\bigr),
\end{gather}
producing the updated BEV query $Q_{\text{PCMF}}$ that enables persona-aware trajectory prediction.

\subsection{Persona-Guided Training Objectives}
Let $\tau_{m,t} \in \mathbb{R}^{2}$ denote the predicted waypoint of persona $m$ at timestep $t$, with $T$ the total number of timesteps. The following two losses leverage the $3 \times 3$ grid structure to provide axis-aligned supervision. \\

\noindent \textbf{Hierarchical Guide Loss.}
Although PCAT can produce anchors appropriate for each persona, naive end-to-end training may fail to enforce the expected physical orderings across the two behavioral axes. We therefore introduce a Hierarchical Guide Loss that enforces axis-aligned orderings using ground-truth margins.
For each persona $m$, we define the trajectory length $l_m = \sum_{t=1}^{T-1} \| \tau_{m,t+1} - \tau_{m,t} \|_2$ and the mean jerk magnitude $j_m$ via third-order finite differences ($\Delta t = 0.5$\,s). Within the same comfort column, higher-urgency trajectories should travel farther; within the same urgency row, lower-comfort trajectories should exhibit higher jerk. Both orderings are enforced with the same margin form:
\begin{gather}
    \mathcal{L}_{\text{urg}} = \frac{1}{|\mathcal{C}_u|} \sum_{(m_h, m_l) \in \mathcal{C}_u} \text{ReLU}\!\Big( \big[l^{\text{gt}}_{m_h} - l^{\text{gt}}_{m_l}\big]_+ - \bigl(l^{\text{pred}}_{m_h} - l^{\text{pred}}_{m_l}\bigr) \Big), \label{eq:loss_urg} \\
    \mathcal{L}_{\text{cmf}} = \frac{1}{|\mathcal{C}_c|} \sum_{(m_l, m_h) \in \mathcal{C}_c} \text{ReLU}\!\Big( \big[j^{\text{gt}}_{m_l} - j^{\text{gt}}_{m_h}\big]_+ - \bigl(j^{\text{pred}}_{m_l} - j^{\text{pred}}_{m_h}\bigr) \Big), \label{eq:loss_cmf}
\end{gather}
where $\mathcal{C}_u$ and $\mathcal{C}_c$ denote the six adjacent urgency pairs (3 columns $\times$ 2) and six adjacent comfort pairs (3 rows $\times$ 2), respectively. The total guide loss $\mathcal{L}_{\text{guide}} = \mathcal{L}_{\text{urg}} + \mathcal{L}_{\text{cmf}}$ yields 12 margin constraints that prevent the model from conflating urgency-driven and comfort-driven trajectory differences. \\

\noindent \textbf{Axis-Decomposed Diversity Loss.}
Applying a single diversity objective over all $M{=}9$ persona pairs risks diagonal mode collapse, where personas differing along both axes still produce similar trajectories. We therefore decompose the loss into intra-axis and inter-axis components.

For the intra-axis loss, we define six groups of three personas sharing one axis (three urgency groups and three comfort groups). Within each group $\mathcal{G}_k$, we compute pairwise average $L_1$ distance matrices from predicted and GT trajectories, convert each row to a temperature-scaled softmax distribution (with diagonal masking), and align the two via symmetric KL divergence:
\begin{gather}
    \mathcal{L}_{\text{KL}}^{(k)} = \frac{1}{2|\mathcal{G}_k|} \sum_{i \in \mathcal{G}_k} \Big[ \text{KL}(q_i \| p_i) + \text{KL}(p_i \| q_i) \Big], \label{eq:loss_kl}
\end{gather}
where $p_i$ and $q_i$ are the softmax distributions derived from predicted and GT distances. A margin term additionally penalizes cases where predicted pairwise distance falls below the GT distance. The intra-axis loss $\mathcal{L}_{\text{Intra}}$ averages both terms over all six groups; the inter-axis loss $\mathcal{L}_{\text{Inter}}$ applies the same KL-margin formulation over all $M{=}9$ personas. The final loss combines both:
\begin{gather}
    \mathcal{L}_{\text{AD}} = \mathcal{L}_{\text{Intra}} + w_{\text{Inter}} \cdot \mathcal{L}_{\text{Inter}}, \label{eq:loss_ad}
\end{gather}
where $w_{\text{Inter}} = 0.2$. The intra-axis term forces diversity among personas differing along one axis, while the inter-axis term regularizes diagonal pairs.

\subsection{Total Loss} 

The total loss function is represented as:
\begin{equation}
    \mathcal{L}_{\text{Total}} = \lambda_1 \mathcal{L}_{\text{traj}} + \lambda_2 \mathcal{L}_{\text{guide}} + \lambda_3 \mathcal{L}_{\text{AD}},
    \label{eq:total_loss}
\end{equation}
where $\mathcal{L}_{\text{traj}}$ is the baseline loss of DiffusionDrive \cite{liao2025diffusiondrive} for trajectory reconstruction and classification, $\mathcal{L}_{\text{guide}}$ is the Hierarchical Guide Loss that enforces axis-aligned physical orderings, and $\mathcal{L}_{\text{AD}}$ is the Axis-Decomposed Diversity Loss that maintains persona separation. $\lambda_1$, $\lambda_2$, and $\lambda_3$ are balancing hyperparameters.

\section{Experiments}
\label{sec:experiments}

\subsection{Evaluation Metrics}
\label{sec:eval_metrics}
We evaluate on NAVSIM \cite{dauner2024navsim}, a closed-loop planning benchmark built on OpenScene \cite{openscene2023}, strictly following its protocol with all simulator settings unchanged. Since the NAVSIM nonreactive simulator is calibrated for normal driving conditions, PDMS may penalize intentionally aggressive or cautious trajectories. We therefore adopt Average Displacement Error (ADE) and Final Displacement Error (FDE) as primary metrics, which directly measure agreement with the persona-specific ground truth. \color{black}We report aggregate PDMS only for comparability with prior work, not as a measure of persona fidelity, as a score calibrated for normal driving does not directly reflect the quality of intentionally non-normal trajectories.\color{black}

\subsection{Implementation Details} 
As our baseline, we adopt DiffusionDrive \cite{liao2025diffusiondrive} with a ResNet-34 backbone \cite{he2016deep}. The input comprises three cropped, downscaled forward-view camera images stitched horizontally into a 1024$\times$256 frame, together with a BEV raster derived from LiDAR. We train from scratch on the navtrain split for 100 epochs using AdamW \cite{loshchilov2017decoupled} (learning rate 0.0006), a global batch size of 128 distributed over 6 NVIDIA RTX A6000 GPUs, and no test-time augmentation. For evaluation on the navtest split, the model outputs an 8-waypoint trajectory over a 4-s horizon. Loss hyperparameters are fixed to $\tau = 0.25$, $\lambda_1 = 10$, $\lambda_2 = \lambda_3 = 1$. The comfort modulation $\alpha_c$ is bounded to $[0.8, 1.2]$. \color{black}Although $\alpha_u$ is left unconstrained, across all 109,269 evaluation samples its observed range is $[-1.42, 1.26]$ ($|\alpha_u|>1$ for only 0.13\%) and final waypoints are hard-clipped during denormalization, so no physically invalid trajectory occurs. Across multiple training seeds, the coefficient of variation of Avg.\ ADE/FDE stays below 0.12\%, and our model wins ADE/FDE in all nine cells. All experiments are implemented in PyTorch \cite{paszke2019pytorch}.\color{black}

\begin{table*}[t]
    \renewcommand{\tabcolsep}{4mm}
    \centering
    \caption{Comparison of closed-loop metrics on the planning-oriented NAVSIM navtest split. We report Avg.\ ADE, Avg.\ FDE, and Avg.\ PDMS averaged across all nine persona categories. \textbf{Bold}/\underline{underlined} fonts indicate the best/second-best results.}
    \resizebox{\linewidth}{!}{
    \begin{tabular}{c c c c}
        \Xhline{3\arrayrulewidth}
    \rule{0pt}{10pt}\bf Methods & \bf Avg. ADE$\downarrow$ & \bf Avg. FDE$\downarrow$ & \bf Avg. PDMS$\uparrow$ \\\hline
        \rule{0pt}{9.0pt}LTF (TPAMI'22) \cite{chitta2022transfuser}  
        &   2.49  &  3.87  &  52.0   \\
        Transfuser (TPAMI'22) \cite{chitta2022transfuser}  
        &   2.53  &  3.96  &  51.9   \\
        UniAD (CVPR'23) \cite{hu2023planning}  
        &   \underline{2.47}  &  \underline{3.83}  &  52.5   \\
        Hydra-MDP (arXiv'24) \cite{li2024hydra}  
        &   6.70  &  11.85 &  41.4   \\
        PARA-Drive (CVPR'24) \cite{Weng2024paradrive}  
        &   4.84  &  9.00  &  51.9   \\
        Traj-LLM (T-IV'24) \cite{lan2024traj}  
        &   2.66  &  4.20  &  54.1   \\
        VisionTRAP (ECCV'24) \cite{moon2024visiontrap}  
        &   5.17  &  9.95  &  50.7   \\
        WoTE (ICCV'25) \cite{li2025end}  
        &   3.35  &  5.71  &  52.4   \\
        DiffusionDrive (CVPR'25) \cite{liao2025diffusiondrive}  
        &   3.93  &  5.79  &  \underline{55.3}   \\
        VADv2 (ICLR'26) \cite{chen2024vadv2}  
        &   6.18  &  10.76 &  47.9   \\\cdashline{1-4}\rule{0pt}{9.0pt}
        \textbf{PersonaDrive}
        &  \bf 2.39  &  \bf 3.71  &  \bf 57.7   \\\Xhline{3\arrayrulewidth}
    \end{tabular}
}
\label{table:main_table}
\vspace{-0.2cm}
\end{table*}

\subsection{Comparison with Previous Methods}
Table \ref{table:main_table} compares ADE and FDE across all nine persona categories. For fair comparison, all baselines share the same frozen text encoder \cite{wang2020minilm}; the text features are concatenated with each method's BEV representation before the prediction head, ensuring identical persona information. Our method achieves the lowest ADE and FDE, reducing ADE by 3.2\% and FDE by 3.1\% over the second-best method, confirming closer fidelity to the persona-specific ground truth. It also attains the highest aggregate PDMS. \color{black}Since PDMS aggregates persona-invariant safety terms with persona-conditional quality terms, we decompose it per-persona in the supplementary material, where the safety terms (NC/DAC/DDC) stay within a consistent floor while the quality terms vary as the persona intends.\color{black}

\subsection{Comparison with One-/Multi-Dimensional}
To quantify the limitation of one-dimensional persona formulations, we construct two proxies from our inference outputs. 1D-Urgency uses the mid-comfort (CM) prediction for all cells in the same urgency row; 1D-Comfort analogously uses the mid-urgency (UM) prediction for each comfort column. These proxies provide a principled upper bound on one-dimensional performance. As shown in Table \ref{table:1d_vs_md}, both proxies incur consistently higher ADE and FDE than the multi-dimensional model. The degradation is largest at against-the-grain cells where urgency and comfort impose opposing demands (\textit{e.g.}, UH--CH: fast yet gentle, UL--CL: slow yet aggressive), confirming that collapsing either axis erases precisely the distinctions that matter most. Neither axis alone is sufficient; the multi-dimensional formulation achieves the lowest error across all nine categories.

\begin{table}[t]
    \centering
    \caption{One-dimensional vs. multi-dimensional persona conditioning on the NAVSIM navtest split. \textbf{Multi-D} is our full model. 1D-Urg. uses the mid-comfort (CM) prediction for all cells in the same urgency row; 1D-Comf. uses the mid-urgency (UM) prediction for all cells in the same comfort column. \textbf{Bold} indicates the best result per cell.}
    \renewcommand{\tabcolsep}{0.5mm}
    \resizebox{\linewidth}{!}{
    \begin{tabular}{c c c c c c c c c c c c}
        \Xhline{3\arrayrulewidth}
        \rule{0pt}{10pt} \bf Method & \bf Metric
        & \bf UH-CL & \bf UH-CM & \bf UH-CH
        & \bf UM-CL & \bf UM-CM & \bf UM-CH
        & \bf UL-CL & \bf UL-CM & \bf UL-CH
        & \bf Avg. \\\hline
        \rule{0pt}{9pt}
        1D-Urg.
        & \multirow{3}{*}{ADE$\downarrow$}
          & 3.30 & \bf 2.96 & 3.07 & 2.76 & \bf 2.35 & 2.35 & 2.73 & \bf 1.93 & 2.56 & 2.67 \\
        1D-Comf.
        & & 4.79 & 4.85 & 4.66 & 2.43 & \bf 2.35 & \bf 2.34 & 5.70 & 6.26 & 7.11 & 4.50 \\\cdashline{1-1}\cdashline{3-12}\rule{0pt}{9pt}
        \bf Multi-D
        & & \bf 2.98 & \bf 2.96 & \bf 2.69 & \bf 2.43 & \bf 2.35 & \bf 2.34 & \bf 2.02 & \bf 1.93 & \bf 1.82 & \bf 2.39 \\
        \hline
        \rule{0pt}{9pt}
        1D-Urg.
        & \multirow{3}{*}{FDE$\downarrow$}
          & 5.81 & \bf 5.15 & 5.42 & 4.14 & \bf 3.52 & 3.53 & 4.06 & \bf 2.63 & 4.21 & 4.27 \\
        1D-Comf.
        & & 11.76 & 11.40 & 10.32 & 3.69 & \bf 3.52 & \bf 3.50 & 9.84 & 10.88 & 12.87 & 8.64 \\\cdashline{1-1}\cdashline{3-12}\rule{0pt}{9pt}
        \bf Multi-D
        & & \bf 5.25 & \bf 5.15 & \bf 4.47 & \bf 3.69 & \bf 3.52 & \bf 3.50 & \bf 2.82 & \bf 2.63 & \bf 2.33 & \bf 3.71 \\
        \Xhline{3\arrayrulewidth}
    \end{tabular}
    }
    \label{table:1d_vs_md}
\end{table}

\begin{table*}[t]
    \renewcommand{\tabcolsep}{0.8mm}
    \renewcommand{\arraystretch}{1.1}
    \centering
    \caption{Effect of our proposed components on the NAVSIM navtest split. All metrics are averaged across all nine persona categories. All variants include the text encoder. Detailed per-persona results are in the supplementary material.}
    \resizebox{0.8\linewidth}{!}{
    \begin{tabular}{c c c c c c c}
        \Xhline{3\arrayrulewidth}
        \bf PCAT & \bf PCMF  & \bf $\mathcal{L}_{AD}$ & \bf Avg. ADE$\downarrow$ & \bf Avg. FDE$\downarrow$ & \bf Avg. PDMS$\uparrow$  \\\hline
        - & - & - & 3.93  &  5.79  &  55.3  \\\cdashline{1-6}
        \cmark & - & - & 3.52 & 5.43 & 56.0  \\
        \cmark & \cmark & - & 3.23 & 4.85 & 56.9  \\
        - & \cmark & \cmark & 2.55 & 3.87 & 57.2  \\
        \cmark & - & \cmark & 2.61 & 3.97 & 57.5  \\\cdashline{1-6}
        \cmark &\cmark &\cmark & \bf 2.39 & \bf 3.71 & \bf 57.7  \\
        \Xhline{3\arrayrulewidth}
    \end{tabular}
    }
    \label{table:ablation}
    \vspace{-0.3cm}
\end{table*}

\subsection{Ablation Study} 
We conduct an ablation study to quantify the effect of the proposed components (Persona-Conditioned Anchor Transform (PCAT), Persona-Conditioned Multi-Modal Fusion (PCMF), axis-decomposed diversity loss $\mathcal{L}_{AD}$). As shown in Table \ref{table:ablation}, all components that incorporate at least one of the proposed modules outperform the text-encoder-only baseline. When all our components are considered, we achieve the highest performance.

\subsection{Discussions} 

\noindent \textbf{Text Encoding Models Comparison.} As shown in Table \ref{table:text_model_comp}, MiniLM achieves the best ADE and FDE despite being the smallest model ($\approx$33M \textit{vs.}\ $\approx$125--140M). Since all variants share the same framework and differ only in the frozen text encoder, the gap reflects embedding quality rather than model capacity. Trained for sentence-level similarity, MiniLM naturally places same-row or same-column persona descriptions close together---the structure PCAT needs for axis decomposition---whereas RoBERTa and DeBERTa, trained with token-level objectives, lack this property. In fact, both models produce higher ADE and FDE than the text-encoder-only baseline (Table \ref{table:ablation}), indicating that token-level embeddings inject noise into the axis decomposition rather than useful persona structure. \color{black}The higher PDMS of RoBERTa does not indicate better persona-aware planning, since PDMS does not measure persona fidelity.\color{black} \\

\begin{table}[t]
    \renewcommand{\tabcolsep}{5.2mm}
    \centering
    \caption{Comparison of different text encoders on the NAVSIM navtest split. We report Avg.\ ADE, Avg.\ FDE, and Avg.\ PDMS averaged across all nine persona categories. Detailed per-persona results are in the supplementary material.}
    \resizebox{0.93\linewidth}{!}{
    \begin{tabular}{cccc}
        \Xhline{3\arrayrulewidth}
    \rule{0pt}{10pt} \bf Text Encoders & \bf Avg. ADE$\downarrow$ & \bf Avg. FDE$\downarrow$ & \bf Avg. PDMS$\uparrow$ \\\hline
        \rule{0pt}{9.0pt}
        DeBERTa \cite{he2020deberta}        & 5.99 & 11.56 & 57.1  \\
        RoBERTa \cite{liu2019roberta}       & 4.72 & 8.76 & \bf 62.0  \\ \rule{0pt}{9.0pt}
        \textbf{MiniLM} \cite{wang2020minilm} & \bf 2.39 & \bf 3.71 & \underline{57.7}  \\
        \Xhline{3\arrayrulewidth}
    \end{tabular}
    }
\label{table:text_model_comp}
\vspace{-0.1cm}
\end{table}

\begin{table}[t]
    \renewcommand{\tabcolsep}{3.5mm}
    \centering
    \caption{Comparison of one-hot persona encoding and text-based persona description on the NAVSIM navtest split. We report Avg.\ ADE, Avg.\ FDE, and Avg.\ PDMS for DiffusionDrive with one-hot labels and our framework with one-hot and text-based persona inputs, averaged across all nine persona categories. More results are in the supplementary material.}
    \resizebox{\linewidth}{!}{
    \begin{tabular}{cccc}
        \Xhline{3\arrayrulewidth}
        \bf Methods & \bf Avg. ADE$\downarrow$ & \bf Avg. FDE$\downarrow$ & \bf Avg. PDMS$\uparrow$ \\\hline
        
        DiffusionDrive (One-Hot) \cite{liao2025diffusiondrive}  
        & 2.67 & 4.20 & 55.6  \\\cdashline{1-4}
        PersonaDrive (One-Hot)
        & 2.54 & 4.01 & 56.6  \\
        \textbf{PersonaDrive (Text)}
        & \bf 2.39 & \bf 3.71 & \bf 57.7  \\
        \Xhline{3\arrayrulewidth}
        \end{tabular}
    }
\label{table:OH_vs_text}
\vspace{-0.3cm}
\end{table}

\noindent \textbf{One-Hot vs.\ Text.}
Table \ref{table:OH_vs_text} reports performance when persona is given as a one-hot vector. Our framework outperforms the baseline even under one-hot inputs, and text-based conditioning further improves over one-hot, with the largest gains in multi-dimensional scenarios where personas share the same urgency but differ in comfort. This supports the argument that text embeddings naturally place same-row or same-column descriptions closer in the embedding space, preserving the similarity structure that multi-dimensional persona control requires, while one-hot vectors are mutually orthogonal regardless of axis sharing. \\

\noindent \textbf{FPS and Model Size.} 
Our method processes all nine persona texts in a single pass during training, increasing training time from 0.636s/batch to 0.853s/batch (34.1\% slower). Inference overhead is minimal: the baseline runs at 0.022s/image (45 FPS) and ours at 0.023s/image (43 FPS), 4.7\% increase. In practical use, persona is specified once before driving, so the text encoding can be computed once and reused. Model size grows from 61.4M to 63.2M parameters (2.75\%).

\begin{figure*}[t] 
  \centering
  \includegraphics[width=\linewidth]{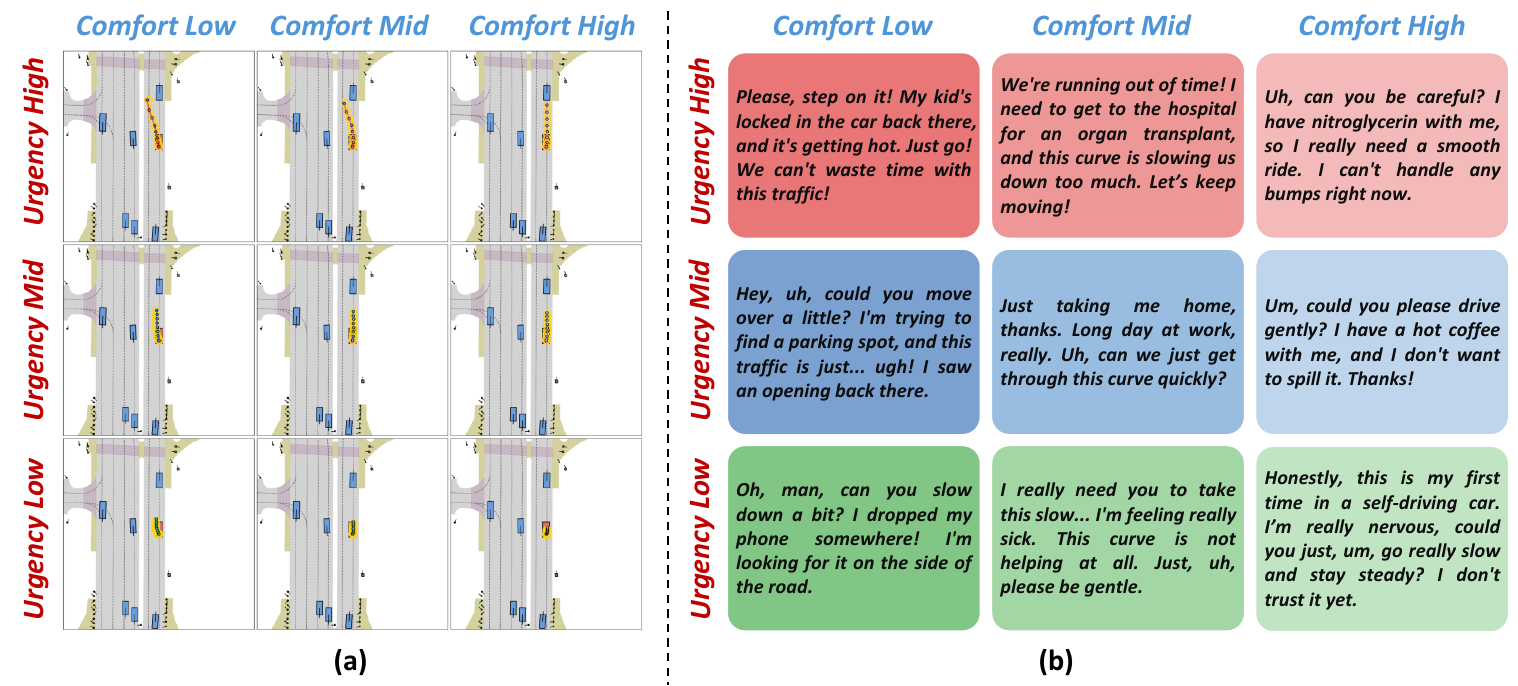}
    \caption{Qualitative results on the NAVSIM dataset. (a) Predicted trajectories under nine personas spanning the Urgency and Comfort axes. (b) Corresponding natural-language persona descriptions used as text input for each cell of the 3×3 grid. Each cell shows the ground-truth persona trajectory (waypoints highlighted in yellow) for one Urgency×Comfort persona.}
  \label{fig:qualitative}
\end{figure*}

\subsection{Visualization Results}
Figure \ref{fig:qualitative} presents qualitative results on NAVSIM scenes. Higher urgency yields longer trajectories with greater forward progress, while higher comfort produces smoother motion. Personas sharing the same urgency but differing in comfort exhibit similar forward progress yet distinct handling profiles, \color{black}remaining within the behavioral envelope spanned by the ground-truth data\color{black}. More examples are in the supplementary material.

\section{Conclusion}
\label{sec:conclusion}
We propose PersonaDrive, a controllable trajectory prediction framework that conditions on natural-language persona descriptions along two orthogonal axes---Temporal Urgency and Ride Comfort---yielding a $3\times3$ grid of nine distinct personas. The PCT dataset pairs each persona with trajectory-text annotations via a four-stage LLM-as-a-judge pipeline, and two dedicated modules---PCAT and PCMF---inject persona cues into anchor priors and BEV features, supervised by a Hierarchical Guide Loss and an Axis-Decomposed Diversity Loss. Experiments results on NAVSIM show consistent improvements across all nine personas, with the largest gains where both axes must be jointly resolved.

\section*{Acknowledgements}
This work was partly supported by IITP-ITRC grant funded by the Korea government (MSIT)(IITP-2026-RS-2023-00258649, 30\%) and partly supported by IITP grant funded by the Korea government (MSIT)(No. RS-2022-II220124, Development of Artificial Intelligence Technology for Self-Improving Competency-Aware Learning Capabilities (50\%), IITP-2023-RS-2023-00266615: Convergence Security Core Talent Training Business Support Program (20\%)).

%
%
\bibliographystyle{splncs04}
\bibliography{main}

\clearpage

\title{PersonaDrive: Controllable Trajectory Prediction with Multi-Dimensional Driving Personas \\ -- \textit{Supplementary Material} --}

\titlerunning{PersonaDrive}

\author{Chan Lee\inst{1}\orcidlink{0009-0004-7827-8579}, Kimin Yun\inst{2}\orcidlink{0000-0002-4493-9437}, Yuseok Bae\inst{2}\orcidlink{0000-0002-4979-2649}, Seong Tae Kim\inst{1}$^{\dagger}$\orcidlink{0000-0002-2132-6021},\\ and Jung Uk Kim\inst{1}$^{\dagger}$\orcidlink{0000-0003-4533-4875}}

\authorrunning{C.~Lee et al.}

\institute{Kyung Hee University, Yong-in, South Korea \\
\email{\{cksdlakstp12, st.kim, ju.kim\}@khu.ac.kr} \and
ETRI, Daejeon, South Korea\\
\email{\{kimin.yun, baeys\}@etri.re.kr}}

\setcounter{page}{1}
\maketitle

\setcounter{section}{0}
\renewcommand\thesection{\arabic{section}}
\setcounter{table}{0}
\renewcommand{\thetable}{S.\arabic{table}}
\setcounter{figure}{0}
\renewcommand{\thefigure}{S.\arabic{figure}}
\setcounter{equation}{0}
\renewcommand{\theequation}{S.\arabic{equation}}

\section{Related Work}

\subsection{End-to-End Autonomous Driving} 
UniAD \cite{hu2023planning} performs detection, tracking, map construction, forecasting, and planning under a shared BEV representation. PARA-Drive \cite{Weng2024paradrive} parallelizes perception, prediction, and planning to increase throughput while maintaining accuracy. TransFuser \cite{chitta2022transfuser} fuses multi-view camera and LiDAR features to improve performance, while LTF \cite{chitta2022transfuser} replaces the LiDAR branch with latent queries, simplifying the model toward a vision-centric formulation. VADv2 \cite{chen2024vadv2} directly predicts distributions over future plans and expands the candidate set to cover more diverse behaviors. Hydra-MDP \cite{li2024hydra} learns a large anchor bank via multi-objective distillation. DiffusionDrive \cite{liao2025diffusiondrive} learns trajectory distributions from a planning-by-diffusion perspective and samples high-diversity candidates. Accordingly, trajectory prediction, which forecasts the future trajectory of the ego vehicle, has become increasingly important in autonomous driving paradigms that directly learn the trajectory plan.

\subsection{LLMs in Driving Tasks} 
Recent autonomous driving research increasingly couples large language models to strengthen instruction following, interaction, and reasoning. LMDrive freezes a pre-trained LLM and, with a multimodal encoder and adapter, generates control signals directly from natural language instructions in a closed-loop end-to-end setting \cite{shao2024lmdrive}. AsyncDriver extracts scene-related textual features asynchronously to assist a conventional planner, which reduces per-step LLM inference latency \cite{chen2024asynchronous}. RDA-Driver improves consistency between explanations and decisions through reasoning–decision alignment, achieving lower collision rates and trajectory prediction error \cite{huang2024making}. On datasets and benchmarks, DriveLM captures dependencies among perception, prediction, and planning via vision–language QA, and defines the GVQA task and metrics to support language-centric driving studies \cite{sima2024drivelm}. DrivingGPT converts the driving world into discrete token sequences and unifies world modeling and planning with an autoregressive transformer \cite{chen2025drivinggpt}. Despite this progress, real-time performance guarantees, safety verification, robustness to distribution shift, and stable multimodal alignment remain open problems. In particular, conditional trajectory generation under explicit user persona and persona-controllable datasets are still limited. We address this gap by representing driving persona as natural language and constructing a persona-conditioned dataset that generates trajectories for nine personas within the same scene, enabling verification of monotonic response to persona changes along both urgency and comfort axes under safety constraints.

\section{Additional Validation for PCT Dataset}

\subsection{Human Evaluation of the PCT Dataset}

\begin{figure*}[t] 
  \centering
  \includegraphics[width=0.92\linewidth]{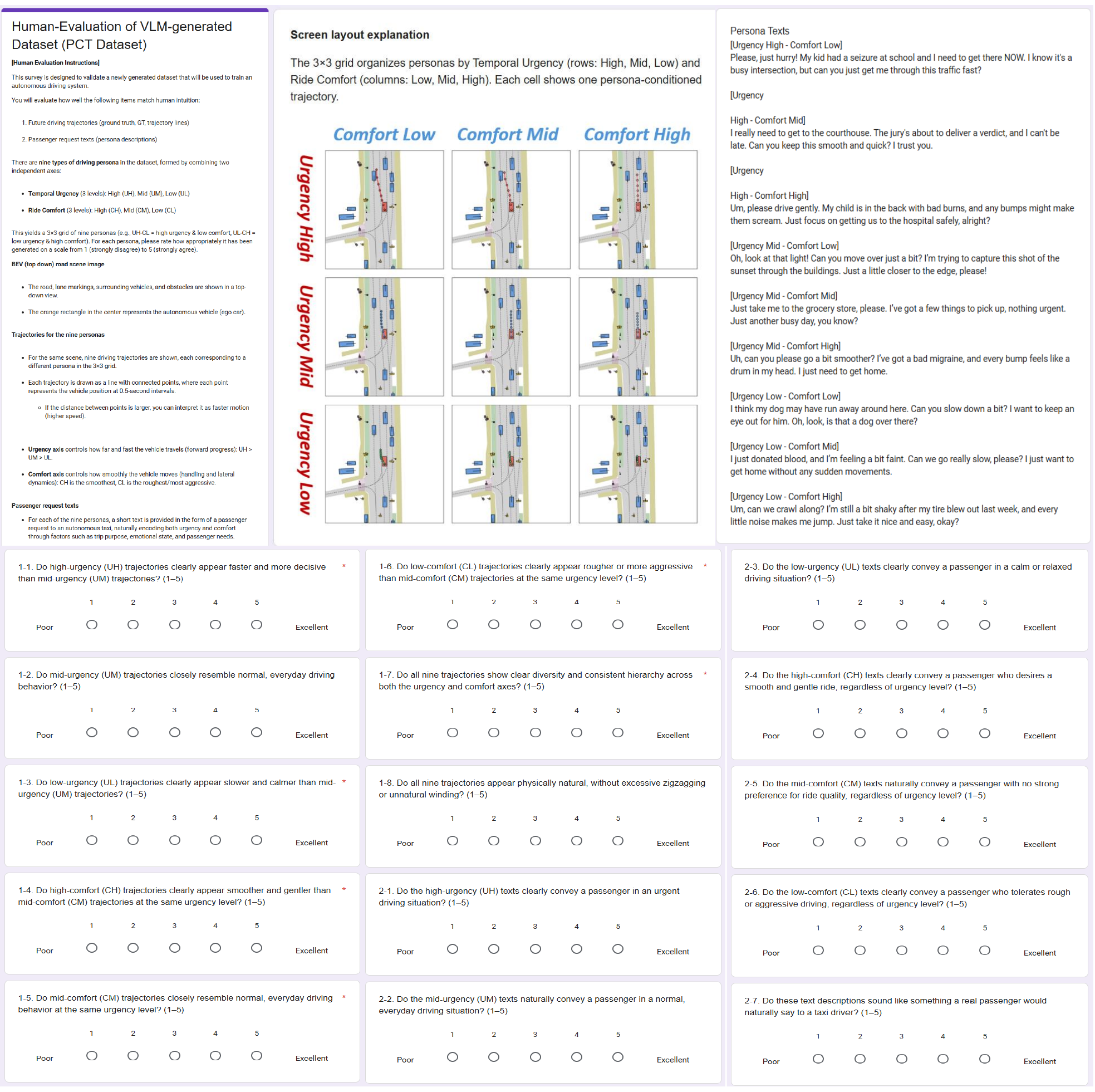}
    \caption{Human evaluation interface for the PCT dataset. For each scene, annotators view a $3\times3$ grid of BEV images, each displaying one persona-conditioned ground-truth trajectory, alongside the corresponding persona descriptions. Annotators rate fifteen Likert-scale items on trajectory quality (urgency ordering, comfort ordering, smoothness, safety, and overall diversity) and text-persona clarity.}
  \label{fig:sup_human_eval}
\end{figure*}

To assess the human plausibility of the LLM-generated annotations in the PCT dataset, we conducted a user study using an online questionnaire (Figure \ref{fig:sup_human_eval}). The goal was to verify that (\textit{i}) the nine ground-truth trajectories are perceived as behaviorally distinct along both the urgency and comfort axes, and (\textit{ii}) the corresponding persona descriptions clearly convey the target persona. \\

\noindent\textbf{Interface.}
Each page of the form showed a $3\times3$ grid of BEV images for a single scene, where each cell corresponds to one persona. Every BEV image displayed the road layout, lane markings, surrounding vehicles, and obstacles in a top-down view, with the ego car highlighted at the center and a single persona-conditioned trajectory rendered as a polyline with points sampled at 0.5s intervals; a larger spacing between points indicates faster motion. Below the grid, we presented nine short persona descriptions arranged in the same $3\times3$ layout, describing the passenger situation, trip purpose, and driving preference. \\

\noindent\textbf{Question Design.}
Participants answered fifteen questions on a five-point Likert scale from 1 (poor) to 5 (excellent):

\begin{itemize}
    \item \textit{GT trajectories (Q1-1 to Q1-8).}
    \begin{itemize}
        \item Q1-1: Do high-urgency (UH) trajectories clearly appear faster and more decisive than mid-urgency (UM) trajectories?
        \item Q1-2: Do mid-urgency (UM) trajectories closely resemble normal, everyday driving behavior?
        \item Q1-3: Do low-urgency (UL) trajectories clearly appear slower and calmer than mid-urgency (UM) trajectories?
        \item Q1-4: Do high-comfort (CH) trajectories clearly appear smoother and gentler than mid-comfort (CM) trajectories at the same urgency level?
        \item Q1-5: Do mid-comfort (CM) trajectories closely resemble normal, everyday driving behavior at the same urgency level?
        \item Q1-6: Do low-comfort (CL) trajectories clearly appear rougher or more aggressive than mid-comfort (CM) trajectories at the same urgency level?
        \item Q1-7: Do all nine trajectories show clear diversity and consistent hierarchy across both the urgency and comfort axes?
        \item Q1-8: Do all nine trajectories appear physically natural, without excessive zigzagging or unnatural winding?
    \end{itemize}
    \item \textit{Texts (Q2-1 to Q2-7).}
    \begin{itemize}
        \item Q2-1: Do the high-urgency (UH) texts clearly convey a passenger in an urgent driving situation?
        \item Q2-2: Do the mid-urgency (UM) texts naturally convey a passenger in a normal, everyday driving situation?
        \item Q2-3: Do the low-urgency (UL) texts clearly convey a passenger in a calm or relaxed driving situation?
        \item Q2-4: Do the high-comfort (CH) texts clearly convey a passenger who desires a smooth and gentle ride, regardless of urgency level?
        \item Q2-5: Do the mid-comfort (CM) texts naturally convey a passenger with no strong preference for ride quality, regardless of urgency level?
        \item Q2-6: Do the low-comfort (CL) texts clearly convey a passenger who tolerates rough or aggressive driving, regardless of urgency level?
        \item Q2-7: Do these text descriptions sound like something a real passenger would naturally say to a taxi driver?
    \end{itemize}
\end{itemize}

\noindent These items jointly evaluate whether the generated trajectories express the expected behavioral signatures along both axes while remaining smooth and safe, and whether the persona descriptions alone communicate the target persona clearly. \\

\begin{table}[t!]
    \renewcommand{\tabcolsep}{3mm}
    \centering
    \caption{Summary of human evaluation scores for the PCT dataset. Ratings are on a five point Likert scale from 1 (poor) to 5 (excellent).}
    \resizebox{0.6\linewidth}{!}{
    \begin{tabular}{c c c c c}
        \Xhline{3\arrayrulewidth}
    \rule{0pt}{10pt} \bf Question & \bf Mean & \bf Std & \bf Median \\\hline
        Q1-1 & 4.84 & 0.46 & 5.0  \\
        Q1-2 & 4.76 & 0.55 & 5.0  \\
        Q1-3 & 4.83 & 0.43 & 5.0  \\
        Q1-4 & 4.64 & 0.50 & 5.0  \\ 
        Q1-5 & 4.64 & 0.62 & 5.0  \\
        Q1-6 & 4.75 & 0.66 & 5.0  \\
        Q1-7 & 4.82 & 0.50 & 5.0  \\
        Q1-8 & 4.61 & 0.65 & 5.0  \\ \cdashline{1-4}
        \rule{0pt}{9.0pt}Q2-1 & 4.64 & 0.62 & 5.0  \\ 
        Q2-2 & 4.49 & 0.73 & 5.0  \\ 
        Q2-3 & 4.63 & 0.58 & 5.0  \\
        Q2-4 & 4.46 & 0.50 & 5.0  \\
        Q2-5 & 4.54 & 0.57 & 5.0  \\
        Q2-6 & 4.59 & 0.75 & 5.0  \\
        \rule{0pt}{9.0pt}Q2-7 & 4.68 & 0.51 & 5.0  \\ \cdashline{1-4}
        Urgency (Q1-1 to Q1-3) & 4.81 & 0.48 & -  \\
        Comfort (Q1-4 to Q1-6) & 4.68 & 0.59 & -  \\
        GT (Q1-1 to Q1-8) & 4.74 & 0.54 & -  \\
        Text (Q2-1 to Q2-7) & 4.57 & 0.61 & -  \\
        All & 4.66 & 0.57 & -  \\\Xhline{3\arrayrulewidth}
    \end{tabular}
}
\label{table:human_eval_stats}
\end{table}

\noindent\textbf{Human Evaluation.}
Each participant evaluated multiple randomly sampled scenes. For every scene, they viewed the $3\times3$ BEV grid, the nine trajectories, and the nine persona descriptions, and then answered fifteen questions (Q1-1 to Q1-8 for trajectory quality and Q2-1 to Q2-7 for text-persona clarity) on a five-point Likert scale. \color{black}We aggregated responses over scenes and participants to obtain item-wise and category-wise statistics. Table \ref{table:human_eval_stats} summarizes the ratings from 21 participants of varying driving experience across all fifteen questions, where enlarging the rater pool from 13 to 21 reduces the standard error of the mean by 21\%\color{black}. Notably, urgency-axis questions (Q1-1 to Q1-3) and comfort-axis questions (Q1-4 to Q1-6) both receive high scores, confirming that annotators perceive behavioral differences along each axis independently. These results indicate that annotators reliably perceive the generated trajectories as realistic, smooth, safe, and consistent with the target persona, and that the persona descriptions alone provide clear cues for both the urgency and comfort dimensions.\\

\begin{figure*}[t] 
  \centering
  \includegraphics[width=0.77\linewidth]{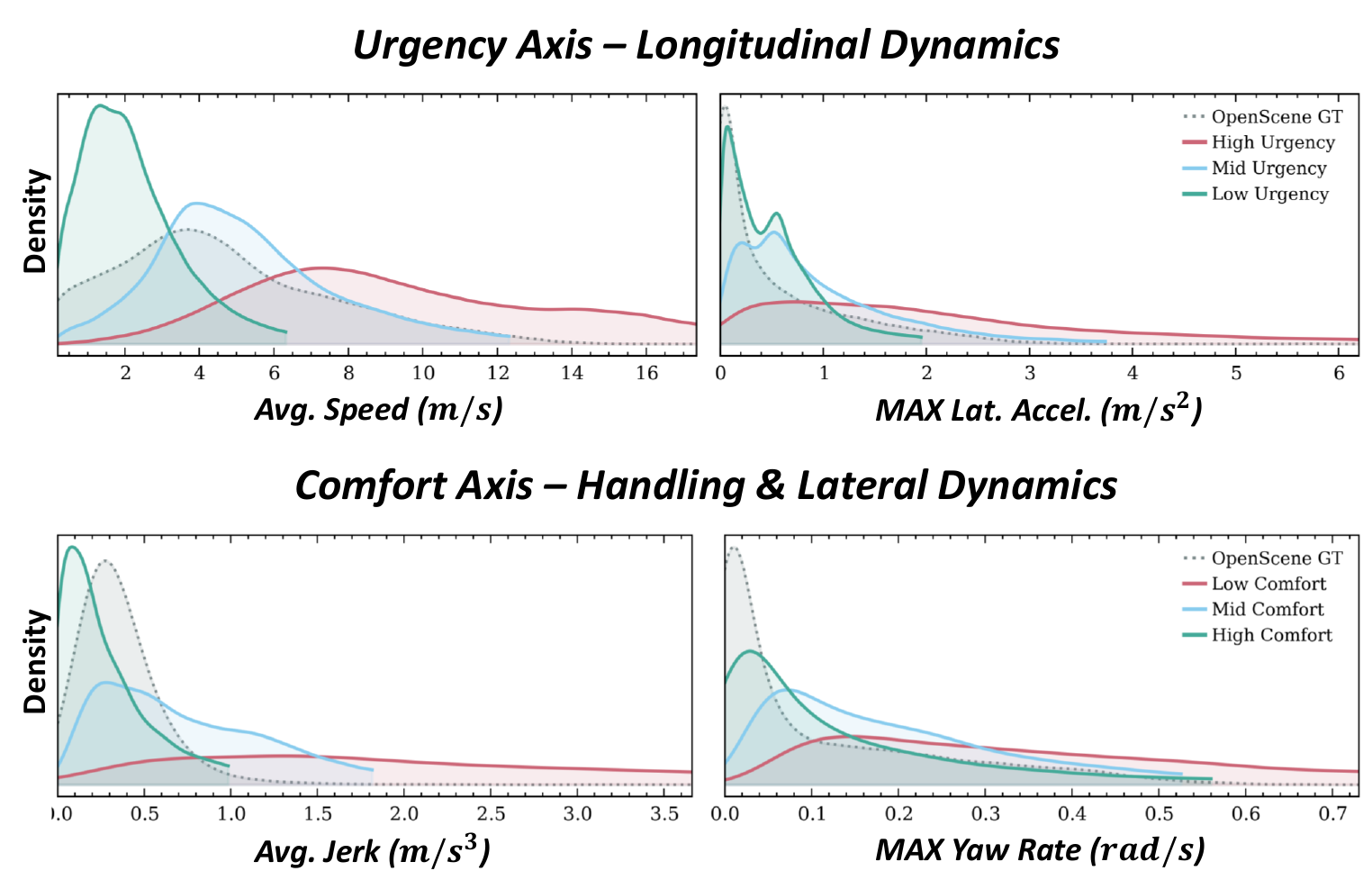}
    \caption{Distribution comparison between OpenScene GT (dotted gray) and PCT persona trajectories. Top: urgency axis (average speed, maximum lateral acceleration). Bottom: comfort axis (average jerk, maximum yaw rate). The OpenScene GT falls between Mid and Low Urgency in longitudinal metrics, and between Mid and High Comfort in handling metrics, confirming that PCT trajectories bracket the real-world driving envelope.}
  \label{fig:sup_statistics}
\end{figure*}

\subsection{Rule-Based Validation Details.}
The rule-based validation stage applies deterministic checks to both trajectories and persona descriptions before passing candidates to the LLM-as-a-Judge.

For trajectories, we enforce five criteria.
First, a \textit{speed ordering} check verifies that the group-wise mean speed strictly decreases from High Urgency to Low Urgency, i.e., $\bar{v}_{\mathrm{UH}} > \bar{v}_{\mathrm{UM}} > \bar{v}_{\mathrm{UL}}$.
Second, a \textit{time-to-collision} (TTC) check requires $\mathrm{TTC} \geq 1.0\text{s}$ with respect to the nearest lead vehicle in the same lane, evaluated under a constant-speed assumption and applied only when the closing speed exceeds $0.5\text{m/s}$.
Third, an \textit{inter-persona diversity} check computes the mean pairwise ADE across all $\binom{9}{2}=36$ trajectory pairs and requires it to be at least $0.3\text{m}$.
Fourth, a \textit{trajectory reversal} check rejects any trajectory whose maximum consecutive heading change exceeds $90^{\circ}$, filtering out physically implausible zigzag patterns.
Fifth, a \textit{direction consistency} check flags trajectories that simultaneously deviate more than $45^{\circ}$ in heading and more than $10\text{m}$ laterally from the ground-truth endpoint, ensuring that all personas follow the same route branch.

For persona descriptions, we verify that every description contains at least 5 characters, that each text falls within 10--200 words, and that the Jaccard similarity between all $\binom{9}{2}=36$ description pairs does not exceed $0.7$, preventing near-duplicate texts across personas.

\subsection{Comparison with Real-World Driving Statistics}
To verify that the PCT dataset produces trajectories within a realistic behavioral envelope, we compare its trajectory statistics against the OpenScene ground-truth distribution. Figure \ref{fig:sup_statistics} overlays kernel density estimates for four metrics grouped by axis.

Along the urgency axis (top row), average speed is clearly separated across urgency levels, and the OpenScene GT peak falls between Mid and Low Urgency. Along the comfort axis (bottom row), average jerk and maximum yaw rate follow the same pattern: Low Comfort produces long-tailed distributions exceeding the OpenScene GT range, while High Comfort is sharply concentrated near zero. In both cases, the OpenScene GT sits in the moderate region, confirming that the PCT generation pipeline produces trajectories that collectively span the full behavioral range observed in real-world driving, while intentionally extending into underrepresented regimes such as emergency and high-comfort scenarios.

\section{Prompt for PCT Dataset Generation}

The PCT dataset is constructed through a four-stage pipeline, summarized in Prompt 1 (Page 22). In Stage 1, GPT-4o-mini receives structured scene context and generates trajectory parameters for all nine personas (Prompt 2 (Page 23)), then generates natural-language persona descriptions grounded in the actual trajectory characteristics (Prompt 3 (Page 24)). Both queries use structured JSON output with strict mode to ensure parseable results. Stages 2 and 3 run validation in parallel: rule-based checks and GPT-4o judge scoring via a trajectory judge (Prompt 4 (Page 25); 9 criteria) and a text judge (Prompt 5 (Page 25); classification + 4 criteria). Failing scenes are regenerated in Stage 4 with failure context injected into the prompt.
Prompt 2 (Page 23) defines the trajectory generation step. The system message assigns the model the role of a driving behavior simulator embodying nine distinct personas and specifies three output parameters per persona: maneuver type, speed factor (relative to GT), and lateral lane offset. Each persona is given a character backstory with target ranges that link urgency to speed and comfort to lateral behavior. Hard constraints enforce a monotonic speed ordering across all nine personas and restrict high-comfort personas to lane-following with zero offset, ensuring that the comfort axis governs trajectory shape independently of the urgency axis.
Prompt 3 (Page 24) defines the persona description generation step. Each persona is mapped to a specific life situation from a curated pool (\textit{e.g.,} UH-CL: desperate crisis, UH-CH: urgent with fragile cargo, UL-CH: genuinely terrified passenger). The model is instructed to write what a real passenger would say to an autonomous taxi in natural spoken English, grounded in the per-persona trajectory statistics so that faster trajectories are paired with urgent situations and slower ones with vulnerable or cautious passengers. A hash-based assignment mechanism ensures that situation types are diversified across scenes.
Prompts 4 and 5 (Page 25) define the LLM-as-a-Judge evaluation used in Stage 3. The trajectory judge (Prompt 4) scores nine blind-shuffled trajectories on criteria including speed level, aggressiveness, smoothness, safety, scene fit, naturalness, traffic rule compliance, maneuver plausibility, and temporal consistency. The text judge (Prompt 5) classifies each description along urgency and comfort axes and scores naturalness, plausibility, scene reactivity, and specificity. Both judges use GPT-4o with blind-shuffled labels to prevent evaluation bias.

\begin{table*}[t]
    \renewcommand{\tabcolsep}{0.5mm}
    \centering
    \caption{Per-persona comparison of closed-loop metrics on the NAVSIM navtest split. We report ADE and FDE for each of the nine persona categories. \textbf{Bold} indicates the best result per cell.}
    \resizebox{\linewidth}{!}{
    \begin{tabular}{c c c c c c c c c c c c c}
        \Xhline{3\arrayrulewidth}
    \rule{0pt}{10pt}\bf Methods & \textbf{Metric}
        & \bf UH-CL & \bf UH-CM & \bf UH-CH
        & \bf UM-CL & \bf UM-CM & \bf UM-CH
        & \bf UL-CL & \bf UL-CM & \bf UL-CH  & \bf Avg. \\\hline
        \rule{0pt}{9.0pt}LTF (TPAMI'22) \cite{chitta2022transfuser} & \multirow{8}{*}{ADE$\downarrow$}
          & 3.01 & 2.99 & 2.78 & 2.62 & 2.47 & 2.47 & 2.10 & 2.03 & 1.92 & 2.49   \\
        Transfuser (TPAMI'22) \cite{chitta2022transfuser}  
        & & 3.09 & 3.05 & 2.80 & 2.66 & 2.52 & 2.53 & 2.17 & 2.05 & 1.93 & 2.53   \\
        UniAD (CVPR'23) \cite{hu2023planning}  
        & & 2.99 & 2.98 & 2.76 & 2.53 & 2.46 & 2.44 & 2.20 & 2.03 & 1.89 & 2.47   \\
        Hydra-MDP (arXiv'24) \cite{li2024hydra}  
        & & 4.83 & 4.75 & 4.57 & 5.11 & 5.42 & 5.70 & 8.97 & 9.86 & 11.11 & 6.70   \\
        PARA-Drive (CVPR'24) \cite{Weng2024paradrive}  
        & & 5.05 & 4.74 & 4.32 & 3.16 & 2.99 & 3.01 & 5.78 & 6.60 & 7.88 & 4.84   \\
        DiffusionDrive (CVPR'25) \cite{liao2025diffusiondrive}  
        & & 4.47 & 4.47 & 4.27 & 4.05 & 3.95 & 3.91 & 3.54 & 3.43 & 3.28 & 3.93   \\
        VADv2 (ICLR'26) \cite{chen2024vadv2}  
        & & 4.85 & 4.65 & 4.40 & 4.64 & 4.90 & 5.12 & 8.03 & 8.89 & 10.10 & 6.18   \\\cdashline{1-1}\cdashline{3-12}\rule{0pt}{9.0pt}
        \textbf{PersonaDrive}
        & & \bf 2.98 & \bf 2.96 & \bf 2.69 & \bf 2.43 & \bf 2.35 & \bf 2.34 & \bf 2.02 & \bf 1.93 & \bf 1.82 & \bf 2.39   \\\Xhline{3\arrayrulewidth}
        \rule{0pt}{9.0pt}LTF (TPAMI'22) \cite{chitta2022transfuser} & \multirow{8}{*}{FDE$\downarrow$}
          & 5.39 & 5.25 & 4.68 & 3.99 & 3.68 & 3.67 & 2.91 & 2.74 & 2.47 & 3.87   \\
        Transfuser (TPAMI'22) \cite{chitta2022transfuser}  
        & & 5.59 & 5.38 & 4.70 & 4.07 & 3.79 & 3.78 & 3.03 & 2.77 & 2.48 & 3.96   \\
        UniAD (CVPR'23) \cite{hu2023planning}  
        & & 5.32 & 5.20 & 4.61 & 3.81 & 3.66 & 3.62 & 3.09 & 2.75 & 2.43 & 3.83   \\
        Hydra-MDP (arXiv'24) \cite{li2024hydra}  
        & & 8.22 & 7.95 & 7.22 & 8.61 & 9.36 & 9.98 & 16.26 & 18.15 & 20.92 & 11.85   \\
        PARA-Drive (CVPR'24) \cite{Weng2024paradrive}  
        & & 11.79 & 10.63 & 9.04 & 4.78 & 4.52 & 4.57 & 9.77 & 11.54 & 14.37 & 9.00   \\
        DiffusionDrive (CVPR'25) \cite{liao2025diffusiondrive}  
        & & 7.57 & 7.41 & 6.77 & 5.82 & 5.61 & 5.54 & 4.73 & 4.53 & 4.17 & 5.79   \\
        VADv2 (ICLR'26) \cite{chen2024vadv2}  
        & & 8.70 & 8.05 & 7.03 & 7.46 & 8.09 & 8.61 & 14.17 & 16.00 & 18.71 & 10.76   \\\cdashline{1-1}\cdashline{3-12}\rule{0pt}{9.0pt}
        \textbf{PersonaDrive}
        & & \bf 5.25 & \bf 5.15 & \bf 4.47 & \bf 3.69 & \bf 3.52 & \bf 3.50 & \bf 2.82 & \bf 2.63 & \bf 2.33 & \bf 3.71   \\\Xhline{3\arrayrulewidth}
    \end{tabular}
}
\label{table:sup_main_table}
\end{table*}

\section{Additional Experiments Details}

\subsection{Architecture Hyperparameters}
The PCAT urgency and comfort branches are each two-layer MLPs with GELU activation and hidden width 128 (256$\to$128$\to$128), projecting $Q_{txt}$ to the urgency scalar $\alpha_u\in\mathbb{R}$ and the comfort vector $\alpha_c\in\mathbb{R}^{T}$. The PCMF module operates at a transformer hidden dimension of 256 with 8 attention heads: the Query Prototype Pool uses $M{=}8$ learnable seed queries, and the Conditional Resampler, Source Gate MLP, and PCMF Fuser share the same 256-dimensional width.

\subsection{Comparison with Previous Methods} 
Table 2 in the main paper reports averaged metrics. Table \ref{table:sup_main_table} extends this with per-persona ADE and FDE across all nine categories, revealing two distinct failure patterns among baselines.

The first group (LTF, Transfuser, UniAD) produces relatively low and uniform ADE across the grid, with errors increasing toward high-urgency cells (\textit{e.g.,} UniAD: UL-CH 1.89 $\to$ UH-CL 2.99). These methods tend toward conservative, moderate-speed trajectories regardless of the persona input, performing well when the GT happens to be slow but underperforming when fast, decisive motion is required.

The second group (Hydra-MDP, PARA-Drive, VADv2) exhibits the opposite asymmetry: errors escalate sharply toward low-urgency cells (e.g., Hydra-MDP: UH-CL 4.83 $\to$ UL-CH 11.11). Because these anchor- or distribution-based methods maintain strong priors toward forward progress, they overshoot when the target persona calls for slow, cautious driving. DiffusionDrive, our base model, is more balanced (ADE range 3.28 to 4.47) but still shows limited sensitivity to persona variation.

PersonaDrive achieves the lowest ADE on every cell and, importantly, maintains a consistent monotonic decrease from UH-CL (2.98) to UL-CH (1.82), indicating that the model correctly modulates both forward progress and trajectory shape according to the target persona rather than defaulting to a fixed behavioral mode.

\color{black}

\begin{table*}[t]
    \renewcommand{\tabcolsep}{1.4mm}
    \renewcommand{\arraystretch}{1.1}
    \centering
    \caption{Per-persona decomposition of PDMS on the NAVSIM navtest split (higher is better). The multiplicative safety gates (NC/DAC/DDC) stay within a bounded floor even at the most against-the-grain cell (UH-CL), so the score does not collapse; the low aggregate PDMS there comes from the quality terms (Comfort, Ego-Progress) and speed-coupled TTC, reflecting an intended trade-off rather than degraded safety.}
    \resizebox{\linewidth}{!}{
    \begin{tabular}{c c c c c c c c c c c}
        \Xhline{3\arrayrulewidth}
        \rule{0pt}{10pt}\bf Metric
        & \bf UH-CL & \bf UH-CM & \bf UH-CH
        & \bf UM-CL & \bf UM-CM & \bf UM-CH
        & \bf UL-CL & \bf UL-CM & \bf UL-CH & \bf Mean \\\hline
        \rule{0pt}{9pt}
        NC$\uparrow$   & 66.9 & 74.6 & 80.6 & 85.3 & 91.2 & 91.8 & 86.3 & 93.1 & 91.0 & 84.5 \\
        DAC$\uparrow$  & 70.7 & 78.0 & 80.4 & 80.3 & 87.1 & 87.6 & 83.8 & 91.2 & 89.9 & 83.2 \\
        TTC$\uparrow$  & 49.9 & 56.9 & 62.4 & 75.1 & 83.7 & 84.9 & 76.0 & 84.3 & 85.0 & 73.1 \\
        Comf.$\uparrow$ & 45.8 & 55.9 & 69.6 & 79.3 & 80.9 & 82.3 & 84.0 & 76.6 & 63.0 & 70.8 \\
        EP$\uparrow$   & 43.8 & 56.1 & 62.7 & 61.0 & 70.1 & 69.5 & 48.9 & 52.4 & 41.9 & 56.3 \\\hline
        \rule{0pt}{9pt}
        \bf PDMS$\uparrow$ & 35.2 & 47.0 & 54.4 & 59.6 & 71.1 & 71.7 & 56.7 & 65.1 & 58.3 & 57.7 \\
        \Xhline{3\arrayrulewidth}
    \end{tabular}
    }
    \label{table:sup_submetric}
\end{table*}

\subsection{PDMS Sub-Metric Decomposition}
PDMS combines multiplicative safety penalties (NC, DAC, DDC) with a weighted average of TTC, Comfort, and Ego-Progress: a violation of the multiplicative terms collapses the whole score, whereas the averaged terms scale it. The former are persona-invariant constraints every persona must satisfy; the latter, especially Comfort and Ego-Progress, are expected to vary with the persona. Table \ref{table:sup_submetric} reports them per persona. At UH-CL, the most against-the-grain cell, the multiplicative safety terms reach their grid-wide minimum (66.9/70.7/73.9) yet remain within a consistent floor, so the score does not collapse; the low aggregate PDMS there (35.2) instead comes from the averaged quality terms (Comfort 45.8, Ego-Progress 43.8 vs.\ 80.9/70.1 at the neutral UM-CM). A drop in aggregate PDMS therefore does not indicate a safety degradation; we report it for comparability (Sec. 4.1) and ground the safety discussion in the persona-invariant terms.

\subsection{Persona Control in Merging and Evasive Scenarios}
We further examine whether persona conditioning remains both controllable and safe in interaction-heavy scenes. On $\sim$1{,}800 merging scenes, the model scales its lane-change rate from $0.94$ at UH-CL to $0.17$ at UL-CH, a $5.5\times$ text-driven range, showing that urgency and comfort jointly modulate maneuver aggressiveness. On $\sim$1,100 evasive scenes, the reaction time spans $0.82$--$1.62$\,s across personas. Throughout, the persona-invariant safety terms (NC/DAC/DDC) remain within the floor reported in Table \ref{table:sup_submetric}, so this behavioral range is achieved without sacrificing safety compliance.
\color{black}

\begin{table*}[t]
    \renewcommand{\tabcolsep}{0.8mm}
    \renewcommand{\arraystretch}{1.1}
    \centering
    \caption{Effect of our proposed components on the NAVSIM navtest split. All metrics are averaged across all nine persona categories. All variants include the text encoder.}
    \resizebox{\linewidth}{!}{
    \begin{tabular}{c c c c c c c c c c c c c c c}
        \Xhline{3\arrayrulewidth}
        \bf Metric & \bf PCAT & \bf PCMF  & \bf $\mathcal{L}_{AD}$ 
        & \bf UH-CL & \bf UH-CM & \bf UH-CH
        & \bf UM-CL & \bf UM-CM & \bf UM-CH
        & \bf UL-CL & \bf UL-CM & \bf UL-CH  & \bf Avg. \\\hline
        \multirow{6}{*}{ADE$\downarrow$}
        & - & - & -              & 4.47 & 4.47 & 4.27 & 4.05 & 3.95 & 3.91 & 3.54 & 3.43 & 3.28 & 3.93  \\\cdashline{2-14}
        & \cmark & - & -         & 4.10 & 4.14 & 3.88 & 3.58 & 3.48 & 3.48 & 3.12 & 3.03 & 2.92 & 3.52  \\
        & \cmark & \cmark & -    & 3.80 & 3.82 & 3.55 & 3.29 & 3.20 & 3.18 & 2.85 & 2.77 & 2.63 & 3.23  \\
        & - & \cmark & \cmark    & 3.12 & 3.11 & 2.86 & 2.63 & 2.54 & 2.52 & 2.17 & 2.14 & 1.95 & 2.55  \\
        & \cmark & - & \cmark    & 3.22 & 3.21 & 2.96 & 2.69 & 2.59 & 2.55 & 2.19 & 2.12 & 1.99 & 2.61 \\\cdashline{2-14}
        & \cmark &\cmark &\cmark & \bf 2.98 & \bf 2.96 & \bf 2.69 & \bf 2.43 & \bf 2.35 & \bf 2.34 & \bf 2.02 & \bf 1.93 & \bf 1.82 & \bf 2.39 \\
        \Xhline{3\arrayrulewidth}
        \multirow{6}{*}{FDE$\downarrow$}
        & - & - & -              & 7.57 & 7.41 & 6.77 & 5.82 & 5.61 & 5.54 & 4.73 & 4.53 & 4.17 & 5.79 \\\cdashline{2-14}
        & \cmark & - & -         & 7.02 & 6.98 & 6.30 & 5.43 & 5.23 & 5.20 & 4.47 & 4.27 & 4.00 & 5.43  \\
        & \cmark & \cmark & -    & 6.41 & 6.36 & 5.68 & 4.84 & 4.65 & 4.61 & 3.94 & 3.77 & 3.42 & 4.85  \\
        & - & \cmark & \cmark    & 5.37 & 5.28 & 4.67 & 3.90 & 3.70 & 3.66 & 2.96 & 2.88 & 2.43 & 3.87 \\
        & \cmark & - & \cmark    & 5.62 & 5.52 & 4.86 & 3.99 & 3.78 & 3.70 & 2.97 & 2.78 & 2.49 & 3.97 \\\cdashline{2-14}
        & \cmark &\cmark &\cmark & \bf 5.25 & \bf 5.15 & \bf 4.47 & \bf 3.69 & \bf 3.52 & \bf 3.50 & \bf 2.82 & \bf 2.63 & \bf 2.33 & \bf 3.71 \\
        \Xhline{3\arrayrulewidth}
    \end{tabular}
    }
    \label{table:sup_ablation}
\end{table*}

\subsection{Ablation Study} 
Table 4 in the main paper reports Avg. ADE, Avg. FDE, and Avg. PDMS averaged across all nine persona categories. While this view is useful to summarize the overall trend of the proposed modules, it does not fully reveal their effect under each persona. To this end, Table \ref{table:sup_ablation} provides a detailed per-persona comparison. As shown, the configuration that includes all proposed components achieves the lowest ADE and FDE across all nine categories. Compared to using only the text encoder, progressively adding PCAT, PCMF, and the diversity loss leads to gradual improvements, with the largest gain observed for the full model. These trends indicate that reshaping anchors into persona-specific priors, aligning visual and textual cues through PCMF, and reducing mode overlap via the diversity loss act in a complementary way to control both forward progress and trajectory shape in a manner consistent with the target persona across both the urgency and comfort axes.

\subsection{Text Encoder Comparison} 
Table \ref{table:sup_text_model_comp} reports per-persona ADE and FDE for different text encoders on the NAVSIM navtest split. MiniLM achieves the lowest ADE and FDE by a substantial margin despite being the smallest model ($\approx$33M vs. $\approx$125 to 140M). Since all variants share the same framework and differ only in the frozen text encoder, the gap reflects embedding quality rather than model capacity. Trained for sentence-level similarity, MiniLM naturally places same-row or same-column persona descriptions close together, which is precisely the structure PCAT needs for axis decomposition, whereas RoBERTa and DeBERTa, trained with token-level objectives, lack this property. In fact, both models produce considerably higher ADE and FDE than even the text-encoder-only baseline (Table \ref{table:sup_ablation}, first row), indicating that token-level embeddings inject noise into the axis decomposition rather than useful persona structure. We therefore adopt MiniLM as the default text encoder.

\begin{table*}[t]
    \renewcommand{\tabcolsep}{1.0mm}
    \centering
    \caption{Comparison of different text encoders on the NAVSIM navtest split, with per-persona ADE and FDE reported across all nine persona categories.}
    \resizebox{\linewidth}{!}{
    \begin{tabular}{c c c c c c c c c c c c}
        \Xhline{3\arrayrulewidth}
    \rule{0pt}{10pt} \bf Models & \bf Metric
        & \bf UH-CL & \bf UH-CM & \bf UH-CH
        & \bf UM-CL & \bf UM-CM & \bf UM-CH
        & \bf UL-CL & \bf UL-CM & \bf UL-CH  & \bf Avg. \\\hline\rule{0pt}{9.0pt} 
        DeBERTa \cite{he2020deberta} & \bf \multirow{3}{*}{ADE$\downarrow$} 
        & 9.27 & 9.10 & 8.60 & 6.22 & 5.86 & 6.52 & 2.19 & 3.13 & 3.02 & 5.99  \\
        RoBERTa \cite{liu2019roberta} &
        & 3.98 & 4.49 & 4.12 & 4.74 & 4.08 & 2.92 & 4.53 & 6.20 & 7.43 & 4.72  \\
        MiniLM \cite{wang2020minilm}  &
        & \bf 2.98 & \bf 2.96 & \bf 2.69 & \bf 2.43 & \bf 2.35 & \bf 2.34 & \bf 2.02 & \bf 1.93 & \bf 1.82 & \bf 2.39  \\\Xhline{3\arrayrulewidth}\rule{0pt}{9.0pt}
        DeBERTa \cite{he2020deberta} & \bf \multirow{3}{*}{FDE$\downarrow$} 
        & 20.78 & 19.52 & 17.61 & 10.76 & 9.97 & 11.56 & 3.32 & 5.39 & 5.11 & 11.56  \\
        RoBERTa \cite{liu2019roberta} &
        & 8.35 & 9.54 & 8.24 & 8.47 & 6.95 & 4.97 & 7.85 & 10.91 & 13.58 & 8.76  \\
        MiniLM \cite{wang2020minilm}  &
        & \bf 5.25 & \bf 5.15 & \bf 4.47 & \bf 3.69 & \bf 3.52 & \bf 3.50 & \bf 2.82 & \bf 2.63 & \bf 2.33 & \bf 3.71  \\\Xhline{3\arrayrulewidth}
    \end{tabular}
   }
\label{table:sup_text_model_comp}
\end{table*}

\begin{table*}[t]
    \renewcommand{\tabcolsep}{0.8mm}
    \centering
    \caption{Comparison of one-hot persona encoding and text-based persona encoding on the NAVSIM navtest split. We report ADE and FDE for DiffusionDrive with one-hot labels and our framework with one-hot and text-based persona inputs, across all nine persona categories.}
    \resizebox{\linewidth}{!}{
    \begin{tabular}{c c c c c c c c c c c c}
        \Xhline{3\arrayrulewidth}
        \rule{0pt}{10pt} \bf Methods & \bf Metric
        & \bf UH-CL & \bf UH-CM & \bf UH-CH
        & \bf UM-CL & \bf UM-CM & \bf UM-CH
        & \bf UL-CL & \bf UL-CM & \bf UL-CH  & \bf Avg. \\\hline \rule{0pt}{9.0pt}
        DiffusionDrive (One-Hot) \cite{liao2025diffusiondrive}  
        & \multirow{3}{*}{ADE$\downarrow$}
          & 3.32 & 3.29 & 3.03 & 2.71 & 2.65 & 2.59 & 2.24 & 2.16 & 2.03 & 2.67  \\
        PersonaDrive (One-Hot)
        & & 3.15 & 3.16 & 2.89 & 2.60 & 2.53 & 2.48 & 2.11 & 2.03 & 1.90 & 2.54 \\\cdashline{1-1}\cdashline{3-12} \rule{0pt}{9.0pt}
        PersonaDrive (Text)
        & & \bf 2.98 & \bf 2.96 & \bf 2.69 & \bf 2.43 & \bf 2.35 & \bf 2.34 & \bf 2.02 & \bf 1.93 & \bf 1.82 & \bf 2.39
        \\ \hline \rule{0pt}{9.0pt}
        DiffusionDrive (One-Hot) \cite{liao2025diffusiondrive}  
        & \multirow{3}{*}{FDE$\downarrow$}
          & 6.02 & 5.87 & 5.18 & 4.15 & 4.00 & 3.88 & 3.13 & 2.94 & 2.61 & 4.20  \\
        PersonaDrive (One-Hot)
        & & 5.66 & 5.60 & 4.87 & 3.98 & 3.86 & 3.76 & 3.01 & 2.83 & 2.52 & 4.01  \\\cdashline{1-1}\cdashline{3-12} \rule{0pt}{9.0pt}
        PersonaDrive (Text)
        & & \bf 5.25 & \bf 5.15 & \bf 4.47 & \bf 3.69 & \bf 3.52 & \bf 3.50 & \bf 2.82 & \bf 2.63 & \bf 2.33 & \bf 3.71
        \\\Xhline{3\arrayrulewidth}
    \end{tabular}
}

\label{table:sup_OH_vs_text}
\end{table*}

\begin{table*}[t]
    \renewcommand{\tabcolsep}{1.0mm}
    \centering
    \caption{Comparison with language-conditioned trajectory prediction methods on the NAVSIM navtest split. All methods use the same frozen MiniLM text encoder and persona descriptions. We report per-persona ADE and FDE across all nine persona categories.}
    \resizebox{\linewidth}{!}{
    \begin{tabular}{c c c c c c c c c c c c}
        \Xhline{3\arrayrulewidth}
        \rule{0pt}{10pt} \bf Methods & \bf Metric
        & \bf UH-CL & \bf UH-CM & \bf UH-CH
        & \bf UM-CL & \bf UM-CM & \bf UM-CH
        & \bf UL-CL & \bf UL-CM & \bf UL-CH  & \bf Avg. \\\hline \rule{0pt}{9.0pt}
        langTraj (ICCV'25) \cite{chang2025langtraj}  
        & \multirow{3}{*}{ADE$\downarrow$}
          & 3.64 & 3.35 & 3.17 & 2.93 & 2.46 & 2.47 & 2.37 & 2.09 & 2.01 & 2.72  \\
        iMotionLLM (WACV'26) \cite{felemban2026imotion}  
        & & 3.44 & 3.51 & 3.33 & 2.97 & 2.94 & 2.88 & 2.37 & 2.29 & 2.15 & 2.87 \\\cdashline{1-1}\cdashline{3-12} \rule{0pt}{9.0pt}
        PersonaDrive
        & & \bf 2.98 & \bf 2.96 & \bf 2.69 & \bf 2.43 & \bf 2.35 & \bf 2.34 & \bf 2.02 & \bf 1.93 & \bf 1.82 & \bf 2.39
        \\ \hline \rule{0pt}{9.0pt}
        langTraj (ICCV'25) \cite{chang2025langtraj}  
        & \multirow{3}{*}{FDE$\downarrow$}
          & 6.70 & 6.50 & 5.99 & 4.83 & 4.07 & 4.04 & 3.45 & 3.07 & 2.95 & 4.66  \\
        iMotionLLM (WACV'26) \cite{felemban2026imotion}  
        & & 6.27 & 6.27 & 5.76 & 4.63 & 4.56 & 4.41 & 3.34 & 3.16 & 2.80 & 4.58  \\\cdashline{1-1}\cdashline{3-12} \rule{0pt}{9.0pt}
        PersonaDrive
        & & \bf 5.25 & \bf 5.15 & \bf 4.47 & \bf 3.69 & \bf 3.52 & \bf 3.50 & \bf 2.82 & \bf 2.63 & \bf 2.33 & \bf 3.71
        \\\Xhline{3\arrayrulewidth}
    \end{tabular}
}

\label{table:sup_lang_baselines}
\end{table*}

\subsection{One-hot vs. Text} 
Table \ref{table:sup_OH_vs_text} compares one-hot persona encoding and text-based persona encoding for DiffusionDrive and our framework. When both use one-hot labels, our method already yields lower ADE and FDE across all nine persona categories than the original DiffusionDrive, which indicates that PCAT, PCMF, and the diversity loss improve persona-conditioned behavior even under coarse categorical supervision. Replacing the one-hot label with a text-based persona description while keeping our planner fixed brings further gains, with the largest improvements in multi-dimensional scenarios where personas share the same urgency but differ in comfort. This supports the argument that text embeddings naturally place same-row or same-column descriptions closer in the embedding space, preserving the similarity structure that multi-dimensional persona control requires, while one-hot vectors are mutually orthogonal regardless of axis sharing.

\subsection{Comparison with Language-Conditioned Baselines}
Table 2 in the main paper evaluates methods that are not designed for controllable trajectory prediction, where text features are appended to each baseline solely to ensure a fair persona-information baseline. Here, we provide a complementary comparison against methods explicitly proposed for language-conditioned trajectory prediction: LangTraj \cite{chang2025langtraj} and iMotionLLM \cite{felemban2026imotion}. Both methods are evaluated on the PCT dataset using the same frozen MiniLM text encoder and persona descriptions, so that differences in performance reflect architectural capacity of each method to resolve multi-dimensional persona conditioning rather than differences in input representation.

As shown in Table \ref{table:sup_lang_baselines}, PersonaDrive consistently outperforms both baselines across all nine persona categories. The gap is largest at against-the-grain cells (\textit{e.g.}, UH-CH, UL-CL), where urgency and comfort impose opposing behavioral demands. This confirms that existing language-conditioned methods, which condition on a single behavioral axis, cannot adequately resolve the two-axis persona space that PersonaDrive is designed to address.

\subsection{One-Dimensional vs. Multi-Dimensional Training}
Table 3 in the main paper compares one-dimensional and multi-dimensional persona conditioning using inference-time proxies. Here, we train separate models under each formulation to confirm that this advantage holds at training time. \\

\noindent\textbf{Training setup.} All models receive the full nine persona texts as input. 1D-Urgency supervises all cells in the same urgency row with the mid-comfort (CM) ground truth, and 1D-Comfort analogously uses the mid-urgency (UM) ground truth. The Multi-D model trains on all nine personas with their corresponding ground truths. \\

\noindent\textbf{Results.} As shown in Table \ref{table:sup_1d_vs_2d}, each 1D model matches Multi-D only at its supervised cells (1D-Urgency at CM column, 1D-Comfort at UM row), and degrades substantially elsewhere. The largest drops occur at against-the-grain cells (\textit{e.g.}, UH-CH, UL-CL), confirming that one-dimensional supervision fails to learn the suppressed axis even when the input text contains the relevant cues. This is consistent with Table 3, confirming the limitation is a training-time phenomenon rather than a proxy artifact. These results reinforce the core motivation of our work: a single urgency spectrum cannot distinguish personas that share the same urgency level but require different ride dynamics, and only multi-dimensional supervision enables the model to resolve both axes jointly.

\begin{table*}[t]
    \renewcommand{\tabcolsep}{1.0mm}
    \centering
    \caption{One-dimensional vs. multi-dimensional training on the NAVSIM navtest split. \textbf{Multi-D} is our full model. 1D-Urg. trains on the CM column only; 1D-Comf. trains on the UM row only; Multi-D trains on all nine personas jointly. We report per-persona ADE and FDE across all nine persona categories.}
    \resizebox{\linewidth}{!}{
    \begin{tabular}{c c c c c c c c c c c c}
        \Xhline{3\arrayrulewidth}
        \rule{0pt}{10pt} \bf Methods & \bf Metric
        & \bf UH-CL & \bf UH-CM & \bf UH-CH
        & \bf UM-CL & \bf UM-CM & \bf UM-CH
        & \bf UL-CL & \bf UL-CM & \bf UL-CH  & \bf Avg. \\\hline 
        \rule{0pt}{9.0pt}1D-Urg.
        & \multirow{3}{*}{ADE$\downarrow$}
          & 3.44 & 3.05 & 3.00 & 2.98 & 2.47 & 2.40 & 2.79 & 1.94 & 2.37 & 2.72  \\
        1D-Comf.
        & & 4.82 & 4.88 & 4.75 & 2.48 & 2.40 & 2.36 & 5.55 & 6.06 & 6.89 & 4.47 \\\cdashline{1-1}\cdashline{3-12} \rule{0pt}{9.0pt}
        Multi-D
        & & \bf 2.98 & \bf 2.96 & \bf 2.69 & \bf 2.43 & \bf 2.35 & \bf 2.34 & \bf 2.02 & \bf 1.93 & \bf 1.82 & \bf 2.39
        \\ \hline 
        \rule{0pt}{9.0pt}1D-Urg.
        & \multirow{3}{*}{FDE$\downarrow$}
          & 6.37 & 5.43 & 5.11 & 4.62 & 3.77 & 3.59 & 4.26 & 2.65 & 3.84 & 4.40  \\
        1D-Comf.
        & & 11.91 & 11.53 & 10.61 & 3.75 & 3.60 & 3.51 & 9.48 & 10.48 & 12.37 & 8.58 \\\cdashline{1-1}\cdashline{3-12} \rule{0pt}{9.0pt}
        Multi-D
        & & \bf 5.25 & \bf 5.15 & \bf 4.47 & \bf 3.69 & \bf 3.52 & \bf 3.50 & \bf 2.82 & \bf 2.63 & \bf 2.33 & \bf 3.71
        \\\Xhline{3\arrayrulewidth}
    \end{tabular}
}
\label{table:sup_1d_vs_2d}
\end{table*}

\color{black}
\subsection{Cross-Dataset Axis Transferability}
To test whether the two-axis decomposition is specific to OpenScene, we apply the PCT generation pipeline unchanged to 1{,}000-scene subsets of nuScenes and Waymo Open Motion. On both datasets the axis structure reproduces: the high-urgency cells exceed the 75th-percentile ego speed of each dataset, the mid-urgency cells match the dataset mean within $\pm1\%$, and the high-to-low urgency progress ratio is $4.31\times$, even larger than the $3.07\times$ observed on OpenScene. This indicates that the axis decomposition is not specific to OpenScene but transfers to other driving datasets.
\color{black}

\begin{table*}[t]
    \renewcommand{\tabcolsep}{1.0mm}
    \centering
    \caption{Effect of PCAT components on the NAVSIM navtest split. We report ADE and FDE for configurations using only the urgency scalar $\alpha_u$, adding the comfort modulation $\alpha_c$, adding the guide loss, and using all three together, across all nine persona categories.}
    \resizebox{\linewidth}{!}{
    \begin{tabular}{c c c c c c c c c c c c c c}
        \Xhline{3\arrayrulewidth}
    \rule{0pt}{10pt} \bf Metric & \bf $\alpha_u$ & \bf $\alpha_c$ & \bf $\mathcal{L}_{\text{Guide}}$ 
        & \bf UH-CL & \bf UH-CM & \bf UH-CH
        & \bf UM-CL & \bf UM-CM & \bf UM-CH
        & \bf UL-CL & \bf UL-CM & \bf UL-CH  & \bf Avg. \\\hline
        \rule{0pt}{9.0pt}\multirow{4}{*}{ADE$\downarrow$}
        & \cmark &      - &      - & 3.28 & 3.25 & 2.97 & 2.80 & 2.64 & 2.63 & 2.26 & 2.20 & 2.05 & 2.68  \\ 
        & \cmark & \cmark &      - & 3.24 & 3.29 & 3.05 & 2.75 & 2.66 & 2.61 & 2.23 & 2.15 & 2.03 & 2.67  \\
        & \cmark &      - & \cmark & 3.02 & 3.05 & 2.79 & 2.49 & 2.41 & 2.40 & 2.05 & 2.01 & 1.88 & 2.46  \\\cdashline{2-14}\rule{0pt}{9.0pt}
        & \cmark & \cmark & \cmark & \bf 2.98 & \bf 2.96 & \bf 2.69 & \bf 2.43 & \bf 2.35 & \bf 2.34 & \bf 2.02 & \bf 1.93 & \bf 1.82 & \bf 2.39
        \\\hline
        \rule{0pt}{9.0pt}\multirow{4}{*}{FDE$\downarrow$}
        & \cmark &      - &      - & 5.80 & 5.68 & 4.96 & 4.25 & 3.95 & 3.94 & 3.16 & 3.00 & 2.65 & 4.15  \\ 
        & \cmark & \cmark &      - & 5.69 & 5.66 & 5.00 & 4.10 & 3.96 & 3.84 & 3.09 & 2.90 & 2.58 & 4.09  \\
        & \cmark &      - & \cmark & 5.35 & 5.33 & 4.66 & 3.78 & 3.62 & 3.58 & 2.86 & 2.70 & 2.41 & 3.81  \\\cdashline{2-14}\rule{0pt}{9.0pt}
        & \cmark & \cmark & \cmark & \bf 5.25 & \bf 5.15 & \bf 4.47 & \bf 3.69 & \bf 3.52 & \bf 3.50 & \bf 2.82 & \bf 2.63 & \bf 2.33 & \bf 3.71
        \\\Xhline{3\arrayrulewidth}
    \end{tabular}
}

\label{table:sup_PCAT_ablation}
\end{table*}

\section{More Detailed Ablation Study on PersonaDrive}

\subsection{Persona-Conditioned Anchor Transform (PCAT)} 
Table \ref{table:sup_PCAT_ablation} summarizes an ablation on the two PCAT modulation terms, $\alpha_u$ (urgency scalar) and $\alpha_c$ (comfort vector), and the guide loss. In all configurations, $\alpha_u$ is estimated from the text and always used, while $\alpha_c$ and the guide loss are added step by step. Using only $\alpha_u$ already yields consistent improvements across all nine persona categories, indicating that modulating the overall trajectory scale is sufficient to form a persona-aware prior that improves upon the original anchors. When the comfort modulation $\alpha_c$ is additionally used, ADE and FDE decrease further, especially for against-the-grain cells (\textit{e.g.,} UH-CH, UL-CL), showing that beyond adjusting trajectory length, PCAT also benefits from per-timestep shape modulation to capture comfort-driven dynamics.

\begin{figure*}[t] 
  \centering
  \includegraphics[width=0.95\linewidth]{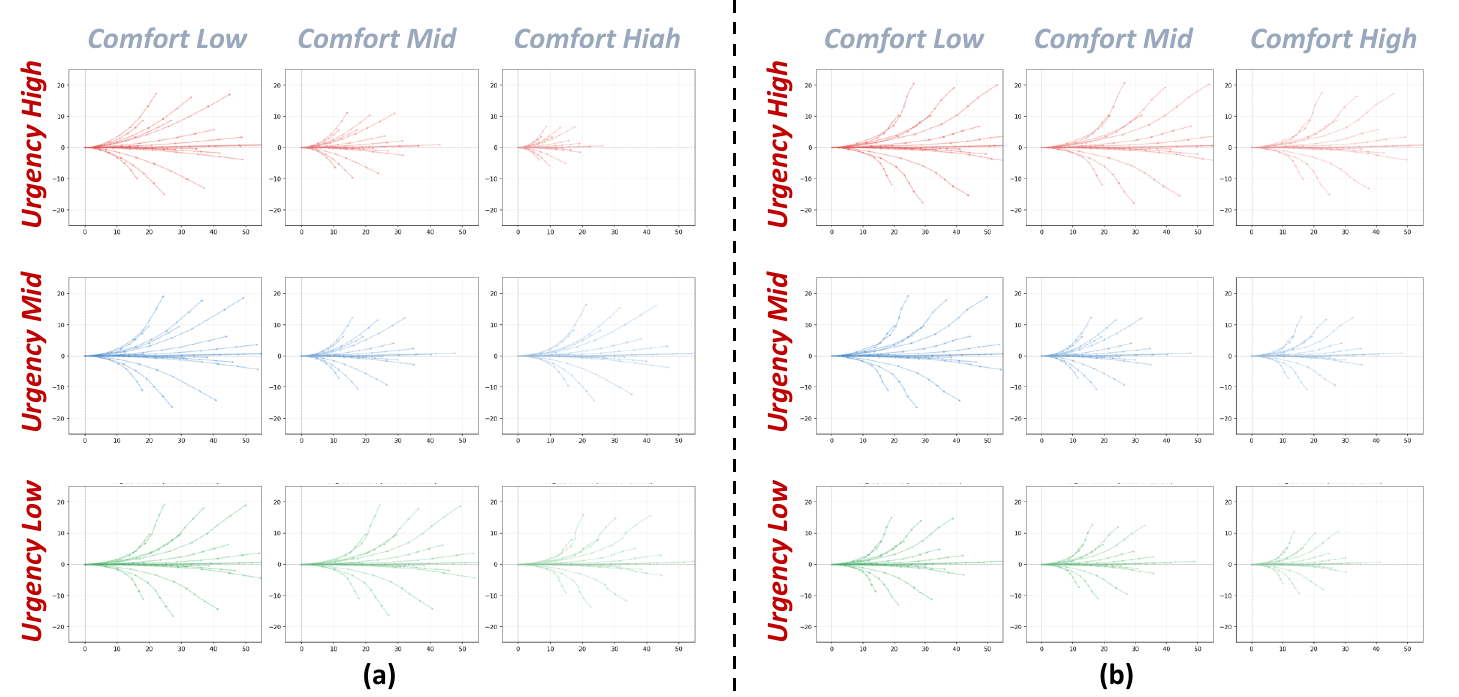}
    \caption{Effect of guide loss on persona-conditioned anchor trajectories. Each column shows anchor sets for three representative personas (high urgency, medium, and low urgency), and rows compare (a) PCAT without guide loss and (b) PCAT with guide loss, where the guide loss enforces a more consistent progress and smoothness ordering across personas.}
  \label{fig:sup_icat_guide}
\end{figure*}

In contrast, configurations with the guide loss encourage the anchor relationships to be aligned with the GT trajectory length and jerk orderings along both axes, which leads to additional gains and reduces cases where the distances between adjacent personas become overly small, thereby helping to keep the axis-aligned separation pattern more stable. As shown in Figure \ref{fig:sup_icat_guide}, without the guide loss (Figure \ref{fig:sup_icat_guide}(a)) some scenes exhibit counter-intuitive scaling, such as a high-urgency anchor becoming shorter than a low-urgency anchor, whereas after adding the guide loss (Figure \ref{fig:sup_icat_guide}(b)) the anchor lengths and shapes consistently follow the progress and smoothness orderings observed in the GT, indicating that the persona inferred from text is more stably reflected in the anchor prior. Finally, when $\alpha_u$, $\alpha_c$, and the guide loss are all used, the model attains the best ADE and FDE across all nine categories, supporting that combining both modulation terms with the Hierarchical Guide Loss yields the most consistent persona-specific anchor priors.

\begin{table*}[t]
    \renewcommand{\tabcolsep}{0.5mm}
    \centering
    \caption{Comparison of text-scene fusion strategies on the NAVSIM navtest split. We report ADE and FDE for Naive Concatenate, Naive Cross-Attention, and the proposed PCMF, across all nine persona categories.}
    \resizebox{\linewidth}{!}{
    \begin{tabular}{c c c c c c c c c c c c}
        \Xhline{3\arrayrulewidth}
        \rule{0pt}{10pt} \bf Methods & \bf Metric 
        & \bf UH-CL & \bf UH-CM & \bf UH-CH
        & \bf UM-CL & \bf UM-CM & \bf UM-CH
        & \bf UL-CL & \bf UL-CM & \bf UL-CH  & \bf Avg. \\\hline
        \rule{0pt}{9.0pt}
        Naive Concatenate
        & \multirow{3}{*}{ADE$\downarrow$}
          & 3.79 & 3.79 & 3.69 & 3.46 & 3.68 & 3.38 & 3.08 & 2.94 & 2.77 & 3.40  \\
        Naive Cross-Attention
        & & 3.18 & 3.18 & 2.90 & 2.65 & 2.59 & 2.56 & 2.20 & 2.08 & 1.93 & 2.59 \\\cdashline{1-1}\cdashline{3-12} \rule{0pt}{9.0pt}
        PCMF
        & & \bf 2.98 & \bf 2.96 & \bf 2.69 & \bf 2.43 & \bf 2.35 & \bf 2.34 & \bf 2.02 & \bf 1.93 & \bf 1.82 & \bf 2.39
        \\\hline\rule{0pt}{9.0pt}
        Naive Concatenate
        & \multirow{3}{*}{FDE$\downarrow$}
          & 6.44 & 6.33 & 5.94 & 5.17 & 5.49 & 4.94 & 4.30 & 4.05 & 3.60 & 5.14  \\
        Naive Cross-Attention
        & & 5.72 & 5.62 & 4.86 & 4.02 & 3.89 & 3.83 & 3.09 & 2.81 & 2.49 & 4.04 \\\cdashline{1-1}\cdashline{3-12} \rule{0pt}{9.0pt}
        PCMF
        & & \bf 5.25 & \bf 5.15 & \bf 4.47 & \bf 3.69 & \bf 3.52 & \bf 3.50 & \bf 2.82 & \bf 2.63 & \bf 2.33 & \bf 3.71
        \\\Xhline{3\arrayrulewidth}
    \end{tabular}
}

\label{table:sup_PCMF_ablation}
\end{table*}

\begin{table*}[t]
    \renewcommand{\tabcolsep}{0.8mm}
    \centering
    \caption{Effect of diversity objective components on the NAVSIM navtest split. We report ADE and FDE for configurations using $\mathcal{L}_{\text{Intra}}$ only, $\mathcal{L}_{\text{Inter}}$ only, and both losses together, across all nine persona categories.}
    \resizebox{\linewidth}{!}{
    \begin{tabular}{c c c c c c c c c c c c c c}
        \Xhline{3\arrayrulewidth}
    \rule{0pt}{10pt} \bf Metric & \bf $\mathcal{L}_{\text{Intra}}$ & \bf $\mathcal{L}_{\text{Inter}}$
        & \bf UH-CL & \bf UH-CM & \bf UH-CH
        & \bf UM-CL & \bf UM-CM & \bf UM-CH
        & \bf UL-CL & \bf UL-CM & \bf UL-CH  & \bf Avg. \\\hline
        \rule{0pt}{9.0pt}
        \multirow{3}{*}{ADE$\downarrow$}
        & \cmark &      - &  3.27 & 3.25 & 3.02 & 2.73 & 2.64 & 2.59 & 2.26 & 2.20 & 2.05 & 2.67 \\\rule{0pt}{9.0pt}
        &      - & \cmark &  3.27 & 3.26 & 2.99 & 2.72 & 2.60 & 2.56 & 2.22 & 2.11 & 2.00 & 2.64  \\\cdashline{2-13}
        & \cmark & \cmark &  \bf 2.98 & \bf 2.96 & \bf 2.69 & \bf 2.43 & \bf 2.35 & \bf 2.34 & \bf 2.02 & \bf 1.93 & \bf 1.82 & \bf 2.39
        \\\hline\rule{0pt}{9.0pt}
        \multirow{3}{*}{FDE$\downarrow$}
        & \cmark &      - &  5.75 & 5.63 & 4.95 & 4.10 & 3.92 & 3.83 & 3.12 & 2.95 & 2.64 & 4.10 \\\rule{0pt}{9.0pt}
        &      - & \cmark &  5.73 & 5.59 & 4.94 & 4.07 & 3.90 & 3.80 & 3.07 & 2.83 & 2.53 & 4.05  \\\cdashline{2-13}
        & \cmark & \cmark &  \bf 5.25 & \bf 5.15 & \bf 4.47 & \bf 3.69 & \bf 3.52 & \bf 3.50 & \bf 2.82 & \bf 2.63 & \bf 2.33 & \bf 3.71
        \\\Xhline{3\arrayrulewidth}
    \end{tabular}
    }
\label{table:sup_loss_ablation}
\end{table*}

\begin{figure*}[t] 
  \centering
  \includegraphics[width=0.98\linewidth]{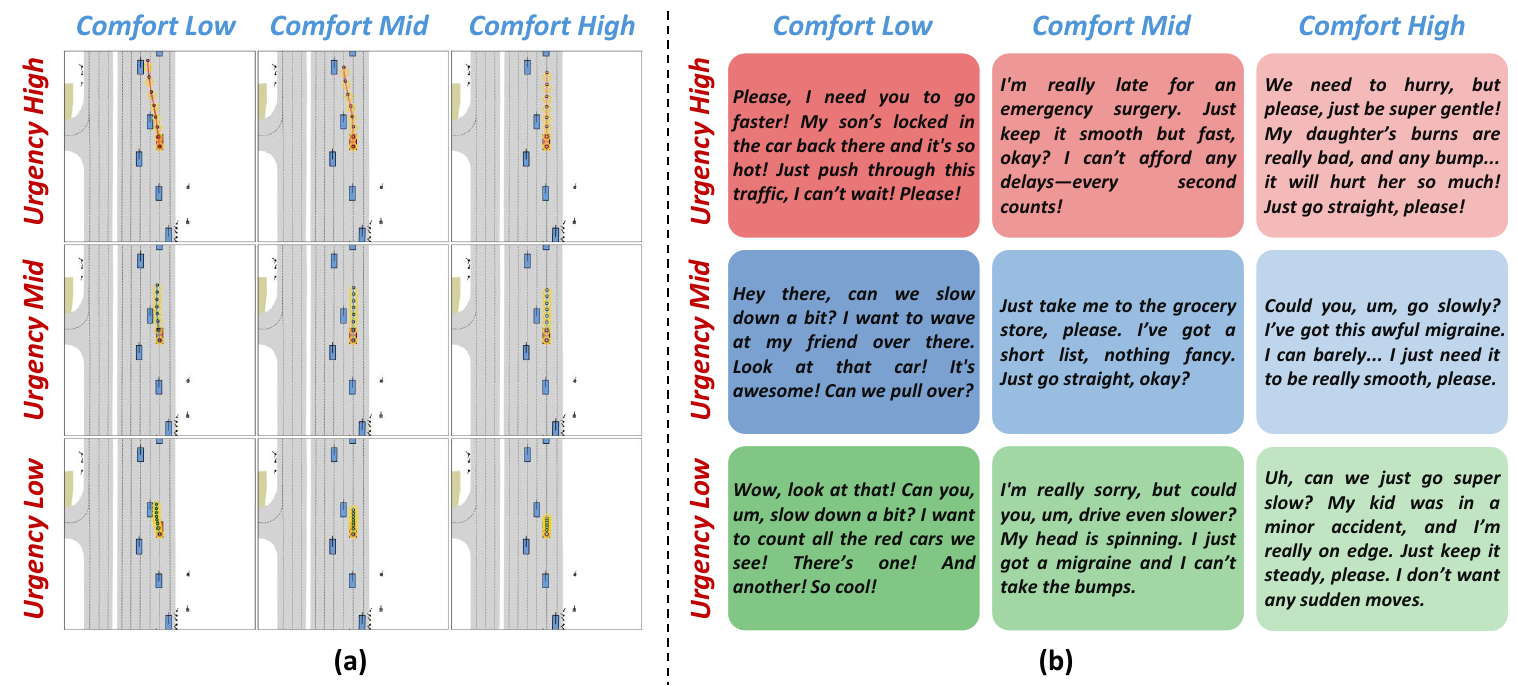}
  \includegraphics[width=0.98\linewidth]{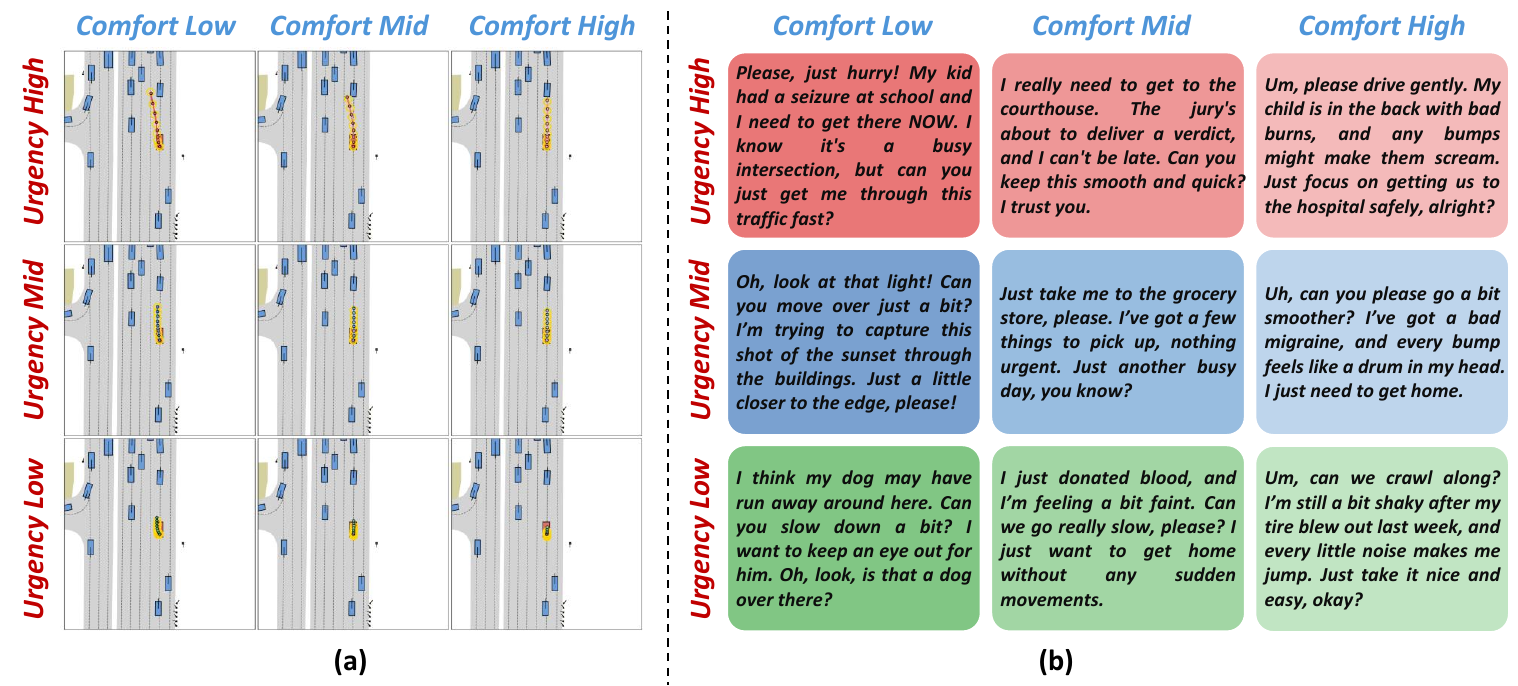}
    \caption{Qualitative results of persona-conditioned trajectory prediction on the NAVSIM dataset. Each row shows one scene, and the $3\times3$ grid indicates predicted trajectories under nine personas spanning Urgency and Comfort. (a) and (b) show the predicted trajectories and the persona text inputs of PersonaDrive, respectively. Each cell shows the ground-truth persona trajectory (waypoints highlighted in yellow) for one Urgency×Comfort persona.}
  \label{fig:sup_qualitative}
\end{figure*}

\subsection{Persona-Conditioned Multi-Modal Fusion (PCMF)} 
Table \ref{table:sup_PCMF_ablation} compares the two naive fusion baselines with the proposed PCMF module. Naive Concatenate simply concatenates the BEV feature and the text feature along the channel dimension and fuses them with a shallow CNN, whereas Naive Cross-Attention passes the BEV query through a sequence of cross-attention blocks over the text queries. Across all nine persona categories, both Naive Concatenate and Naive Cross-Attention provide reasonably stable performance, but PCMF achieves the lowest ADE and FDE. This suggests that naive fusion tends to mix the text signal relatively uniformly over the entire BEV representation and dilute persona information, while PCMF explicitly aligns persona cues with the BEV queries through query-conditioned compression and gated fusion, enabling more effective modeling of the target persona and leading to more stable improvements in both forward progress and trajectory shape across all nine categories.

\subsection{Axis-Decomposed Diversity Loss: $\mathcal{L}_{\text{Intra}}$ and $\mathcal{L}_{\text{Inter}}$} 
Table \ref{table:sup_loss_ablation} summarizes the ablation results for the two loss terms that compose the Axis-Decomposed Diversity Loss, $\mathcal{L}_{\text{Intra}}$ and $\mathcal{L}_{\text{Inter}}$, across all nine persona categories. Using only $\mathcal{L}_{\text{Intra}}$ or only $\mathcal{L}_{\text{Inter}}$ keeps ADE and FDE at reasonable levels, but the configuration that uses both losses achieves the lowest errors. This is consistent with the design: $\mathcal{L}_{\text{Intra}}$ aligns the relative separation pattern among personas sharing one axis (three urgency groups and three comfort groups) to the GT via symmetric KL divergence and margin penalties, securing intra-axis diversity in a way that is robust to scene-wise scale changes. $\mathcal{L}_{\text{Inter}}$ applies the same formulation over all nine personas, regularizing diagonal pairs that differ along both axes and thereby suppressing diagonal mode collapse. In other words, using $\mathcal{L}_{\text{Intra}}$ and $\mathcal{L}_{\text{Inter}}$ together allows the predicted distribution to preserve the relative structure of the GT while keeping sufficient separation between personas along both axes, which leads to more stable improvements across all nine categories.

\begin{figure*}[t] 
  \centering
  \includegraphics[width=0.75\linewidth]{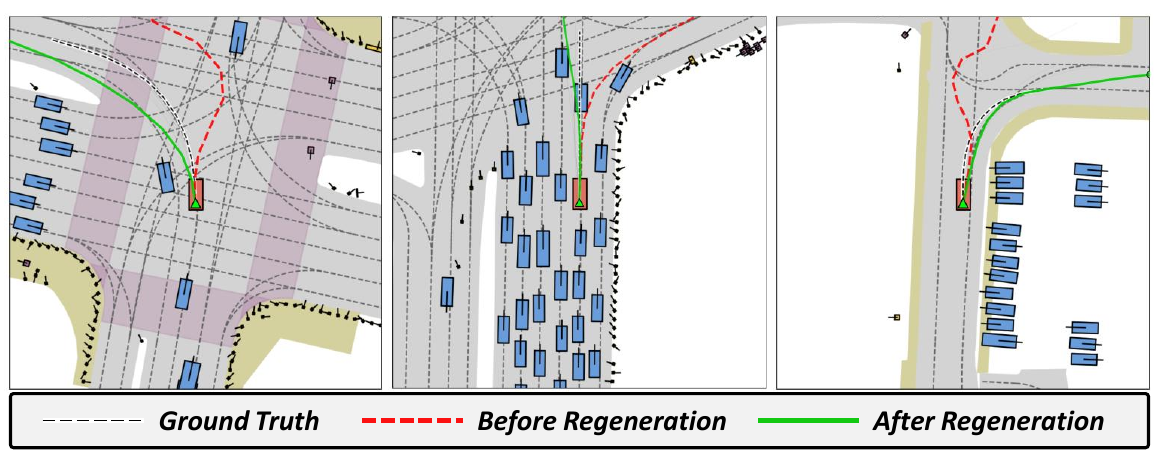}
    \caption{Representative failure cases and their recovered results. Each scene overlays the ground-truth trajectory (dashed gray), the initial failed generation (dashed red), and the recovered trajectory after failure-aware regeneration (solid green), where the specific failure reason is injected into the regeneration prompt. \color{black}Over the full OpenScene generation, 31.4\% of scenes were flagged for rule-based and 4.7\% for quality-based (LLM-judge) regeneration; after up to three retries, 0.18\% remained unrecovered and fell back to the best-scoring candidate.\color{black}}
  \label{fig:sup_failure}
\end{figure*}

\section{Additional Qualitative Results of PersonaDrive}
Figure \ref{fig:sup_qualitative} presents additional qualitative examples that complement Figure 6 in the main paper. Across both scenes, the predicted trajectories faithfully reflect the two-axis persona structure: higher urgency yields longer trajectories with greater forward progress, while higher comfort produces smoother, gentler motion. Personas sharing the same urgency row but differing in comfort exhibit similar travel distance yet visibly distinct handling profiles, and vice versa. The results confirm that PersonaDrive maintains clear persona separation across diverse scene layouts without collapsing to a single default behavior.

\section{Limitation and Failure Case Analysis}

Our persona space is currently defined as a discrete $3\times3$ grid, and extending it to a continuous manifold is a promising future direction. The current evaluation relies on the NAVSIM nonreactive simulator calibrated for normal driving, and evaluating persona-conditioned planning in a fully reactive setting remains an open challenge. Additionally, real-world personas may involve factors beyond urgency and comfort, such as route familiarity or weather adaptation.

Since the PCT dataset is constructed via LLM-based generation, complex scenes such as multi-lane intersections with dense surrounding traffic can cause the LLM to produce trajectories that violate validation constraints. Figure \ref{fig:sup_failure} shows representative failure cases (left) and their recovered results after failure-aware regeneration (right). While LLM-based generation has inherent limitations in spatially complex scenes, the closed-loop regeneration pipeline effectively recovers the majority of these cases.

\color{black}
\section{Continuous Interpolation Across Persona Cells}
\begin{figure}[t]
    \centering
    \includegraphics[width=0.999\linewidth]{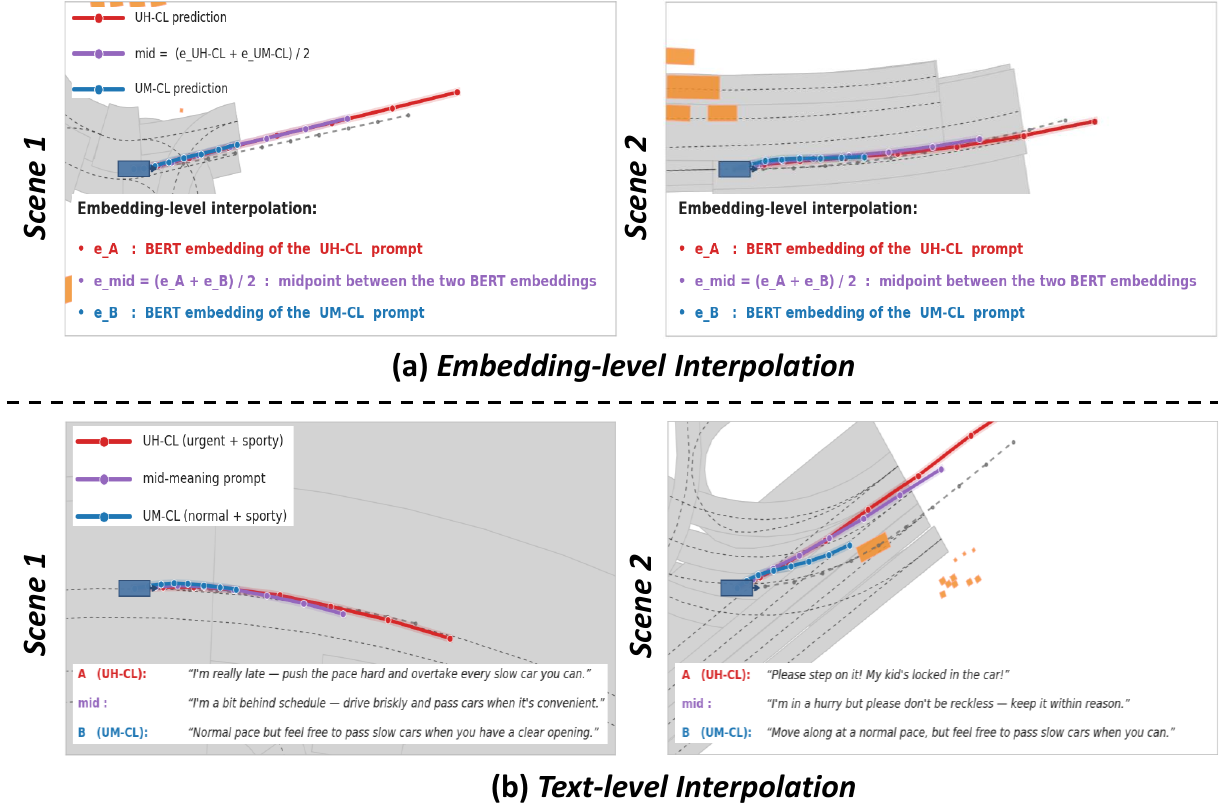}
    \caption{Interpolation between UH-CL and UM-CL: (a) between cell embeddings, (b) between cell-defining prompts. Intermediate trajectories lie smoothly between the two cell predictions, showing that the model treats text as a continuous control signal rather than nine discrete labels.}
    \label{fig:interp}
\end{figure}

The $3\times3$ grid structures training supervision only, not the inference space. Because the frozen text encoder maps any prompt to a continuous embedding and PCAT modulates $\alpha_u, \alpha_c$ continuously, the model reads continuous meaning from text rather than collapsing prompts to nine fixed categories. Figure \ref{fig:interp} demonstrates this: interpolating (a) between two cell embeddings and (b) between two cell-defining prompts both produce a smooth trajectory traversal, where the intermediate predictions lie between the two endpoint-cell trajectories.

\color{black}

\begin{table}[t!]
    \renewcommand{\tabcolsep}{4mm}
    \centering
    \caption{Cross-family LLM-judge agreement with PCT cell labels on 1{,}000 samples (\%). We report top-1 cell accuracy, Cohen's $\kappa$, and per-axis accuracy.}
    \resizebox{0.8\linewidth}{!}{
    \begin{tabular}{l c c c c}
        \Xhline{3\arrayrulewidth}
    \rule{0pt}{10pt}\bf Judge & \bf Top-1 & \bf $\kappa$ & \bf Urgency & \bf Comfort \\\hline
        \rule{0pt}{9.0pt}Gemini Flash & 82.6 & 0.804 & 90.2 & 90.6 \\
        Claude Opus  & \textbf{98.0} & \textbf{0.978} & \textbf{100.0} & \textbf{98.0} \\
        \Xhline{3\arrayrulewidth}
    \end{tabular}
    }
\label{table:sup_crossjudge}
\end{table}

\color{black}
\subsection{Cross-Family LLM-Judge Agreement}
To test whether the PCT cell labels are specific to the GPT-family generator/judge, we re-judge 1{,}000 randomly sampled scenes with two independent model families. As shown in Table~\ref{table:sup_crossjudge}, both judges show substantial-to-near-perfect agreement with the PCT labels (Claude Opus $\kappa=0.978$, Gemini Flash $\kappa=0.804$), confirming that the persona assignments are not GPT-family artifacts.

\subsection{Semantic Grounding and Robustness to Surface Form}
A structured $3\times3$ grid raises the concern that the model may map recurring phrases to nine latent classes rather than read language. Three experiments indicate genuine grounding.

\noindent\textbf{Keyword-Stripped Prompts.} Removing axis-revealing keywords (\textit{e.g.,} ``urgent'', ``gentle'', ``careful'') from 90 prompts still yields top-1 cell accuracy of $32.2\%$ ($2.9\times$ the $11.1\%$ nine-way chance) and urgency-axis accuracy of $60.0\%$ ($1.8\times$ the $33.3\%$ chance), showing the model infers persona from context rather than surface keywords.

\noindent\textbf{Unseen free-form prompts.} On 27 free-form requests written by nine lab members, top-1 accuracy is $44.4\%$ ($4\times$ chance) and at-least-one-axis accuracy is $85.2\%$; all four full misses are one grid step away, indicating graceful degradation rather than collapse.

\noindent\textbf{Paraphrase robustness.} Across 45 paraphrased PCT prompts, the predicted cell is unchanged $88.9\%$ of the time, and the median shift in the urgency scalar $\alpha_u$ is $0.156$---smaller than the within-cell standard deviation ($0.12$--$0.31$)---showing that semantically equivalent rewordings are mapped consistently rather than to different categories.

\begin{table}[t!]
    \renewcommand{\tabcolsep}{4mm}
    \centering
    \caption{Loss-related hyperparameter sweep on the NAVSIM navtest split. Default settings are marked with $^{*}$. All three sit in a smooth basin.}
    \resizebox{0.9\linewidth}{!}{
    \begin{tabular}{l c c c}
        \Xhline{3\arrayrulewidth}
    \rule{0pt}{10pt}\bf Configuration & \bf Avg. ADE$\downarrow$ & \bf Avg. FDE$\downarrow$ & \bf Avg. PDMS$\uparrow$ \\\hline
        \multicolumn{4}{l}{\rule{0pt}{9.0pt}\textit{(a) KL temperature} $\tau$} \\
        $\tau=0.10$        & 2.892 & 4.406 & 56.3 \\
        $\tau=0.25^{*}$    & \textbf{2.391} & \textbf{3.706} & 57.7 \\
        $\tau=0.50$        & 2.759 & 4.265 & 56.8 \\
        $\tau=1.00$        & 2.772 & 4.274 & 57.4 \\ \cdashline{1-4}
        \multicolumn{4}{l}{\rule{0pt}{9.0pt}\textit{(b) Inter-axis weight} $w_{\text{Inter}}$} \\
        $w=0.05$           & 2.678 & 4.183 & 56.4 \\
        $w=0.20^{*}$       & \textbf{2.391} & \textbf{3.706} & 57.7 \\
        $w=0.50$           & 2.812 & 4.327 & 57.0 \\
        $w=1.00$           & 2.694 & 4.191 & 56.2 \\ \cdashline{1-4}
        \multicolumn{4}{l}{\rule{0pt}{9.0pt}\textit{(c) Comfort bound} $(\gamma_c,\Delta_c)$} \\
        $(0.7,0.6)$ wide   & 2.776 & 4.263 & \textbf{58.6} \\
        $(0.8,0.4)^{*}$    & \textbf{2.391} & \textbf{3.706} & 57.7 \\
        $(0.9,0.2)$ narrow & 2.798 & 4.318 & 56.5 \\
        \Xhline{3\arrayrulewidth}
    \end{tabular}
    }
\label{table:sup_loss_sweep}
\end{table}

\subsection{Hyperparameter Sensitivity and Comfort-Bound Choice}
Table \ref{table:sup_loss_sweep} sweeps the three loss-related hyperparameters on the full navtest split. All three lie in a smooth basin around the default, so the method is not fragile to these choices. The comfort bound $(\gamma_c,\Delta_c)=(0.8,0.4)$, i.e. $\alpha_c\in[0.8,1.2]$, matches the OpenScene mean-jerk ratio of $1.15\pm0.07$, motivating the chosen range rather than setting it arbitrarily. Across multiple training seeds, the coefficient of variation of Avg.\ ADE/FDE stays $\leq 0.12\%$, and our model wins ADE/FDE in all nine cells.

\color{black}

\section{More Details of Persona-Guided Training Objectives}

The nine personas form a $3 \times 3$ grid of three urgency levels (High, Medium, Low) $\times$ three comfort levels (Low, Medium, High), so that each column shares the same comfort level and each row shares the same urgency level. Let $\tau_{m,t} \in \mathbb{R}^2$ denote the predicted waypoint of persona $m$ at timestep $t$, with $T$ the total number of timesteps. The following two losses leverage this grid structure to provide axis-aligned supervision. \\

\noindent \textbf{Hierarchical Guide Loss.}
Although PCAT can produce anchors appropriate for each persona, naive end-to-end training may fail to enforce the expected physical orderings across the two behavioral axes. To address this, we introduce a Hierarchical Guide Loss that explicitly enforces axis-aligned orderings among the nine persona trajectories using ground-truth margins.

For the \textit{urgency axis}, within the same comfort column, trajectories with higher urgency should travel farther. We compute the trajectory length for each persona $m$ as the sum of segment-wise L2 norms:
\begin{equation}
    l_m = \sum_{t=1}^{T-1} \left\| \tau_{m,t+1} - \tau_{m,t} \right\|_2,
    \label{eq:traj_length}
\end{equation}
The urgency guide loss enforces length ordering along each of the three comfort columns:
\begin{equation}
    \mathcal{L}_{\text{urg}} = \frac{1}{|\mathcal{C}_u|} \sum_{(m_h, m_l) \in \mathcal{C}_u} \text{ReLU}\!\Big( \big[l^{\text{gt}}_{m_h} - l^{\text{gt}}_{m_l}\big]_+ - \bigl(l^{\text{pred}}_{m_h} - l^{\text{pred}}_{m_l}\bigr) \Big),
    \label{eq:loss_urg}
\end{equation}
where $[\cdot]_+ = \max(0, \cdot)$, and $\mathcal{C}_u$ denotes the set of six adjacent urgency pairs across the three comfort columns (3 columns $\times$ 2 adjacent pairs). The clamped GT margin ensures that when the GT itself violates the expected ordering due to noise, the margin is set to zero rather than producing a misleading gradient.

For the \textit{comfort axis}, within the same urgency row, trajectories with lower comfort sensitivity should exhibit higher jerk (\textit{i.e.,} more abrupt motion). We compute the mean jerk magnitude via third-order finite differences ($\Delta t = 0.5$\,s):
\begin{equation}
    j_m = \frac{1}{T-3} \sum_{t=1}^{T-3} \left\| \frac{\tau_{m,t+3} - 3\,\tau_{m,t+2} + 3\,\tau_{m,t+1} - \tau_{m,t}}{\Delta t^3} \right\|_2,
    \label{eq:traj_jerk}
\end{equation}
The comfort guide loss enforces jerk ordering along each of the three urgency rows:
\begin{equation}
    \mathcal{L}_{\text{cmf}} = \frac{1}{|\mathcal{C}_c|} \sum_{(m_l, m_h) \in \mathcal{C}_c} \text{ReLU}\!\Big( \big[j^{\text{gt}}_{m_l} - j^{\text{gt}}_{m_h}\big]_+ - \bigl(j^{\text{pred}}_{m_l} - j^{\text{pred}}_{m_h}\bigr) \Big),
    \label{eq:loss_cmf}
\end{equation}
where $\mathcal{C}_c$ denotes the set of six adjacent comfort pairs across the three urgency rows (3 rows $\times$ 2 adjacent pairs), and $(m_l, m_h)$ denotes a pair in which $m_l$ has lower comfort sensitivity (expected higher jerk) than $m_h$. The total guide loss combines both axes:
\begin{equation}
    \mathcal{L}_{\text{guide}} = \mathcal{L}_{\text{urg}} + \mathcal{L}_{\text{cmf}},
    \label{eq:loss_guide}
\end{equation}
yielding 12 margin constraints in total (6 urgency + 6 comfort), which provide explicit axis-aligned supervision that prevents the model from conflating urgency-driven and comfort-driven trajectory differences. \\ 

\noindent \textbf{Axis-Decomposed Diversity Loss.}
Applying a single diversity objective over all $M{=}9$ persona pairs risks mixing signals from different behavioral axes, leading to diagonal mode collapse where personas that differ along both axes may still produce similar trajectories. To address this, we decompose the diversity loss into \textit{intra-axis} and \textit{inter-axis} components.

For the \textit{intra-axis} loss, we define six groups of three personas, each sharing one axis and varying along the other: the three columns form the \textit{urgency groups} (fixing comfort, varying urgency) and the three rows form the \textit{comfort groups} (fixing urgency, varying comfort).

For each group $\mathcal{G}_k$ ($|\mathcal{G}_k|{=}3$), we compute the pairwise average L1 distance matrices from predicted and GT trajectories:
\begin{gather}
    D^{\text{pred}}(i,j) = \frac{1}{T} \sum_{t=1}^{T} \left\| \hat{Y}_i(t) - \hat{Y}_j(t) \right\|_1, \label{eq:dpred} \\
    D^{\text{GT}}(i,j) = \frac{1}{T} \sum_{t=1}^{T} \left\| Y_i(t) - Y_j(t) \right\|_1, \label{eq:dgt}
\end{gather}
where $i, j \in \mathcal{G}_k$. Each row is converted to a probability distribution via temperature-scaled softmax (with diagonal masking), and the predicted structure is aligned to the GT through symmetric KL divergence:
\begin{equation}
    \mathcal{L}_{\text{KL}}^{(k)} = \frac{1}{2|\mathcal{G}_k|} \sum_{i \in \mathcal{G}_k} \Big[ \text{KL}(q_i \| p_i) + \text{KL}(p_i \| q_i) \Big],
    \label{eq:loss_kl}
\end{equation}
where $p_i$ and $q_i$ are the softmax distributions derived from the predicted and GT distances, respectively. A margin term additionally penalizes cases where the predicted pairwise distance falls below the GT distance:
\begin{equation}
    \mathcal{L}_{\text{margin}}^{(k)} = \frac{1}{|\mathcal{U}_k|} \sum_{(i,j) \in \mathcal{U}_k} \text{ReLU}\!\bigl(D^{\text{GT}}(i,j) - D^{\text{pred}}(i,j)\bigr),
    \label{eq:loss_margin}
\end{equation}
where $\mathcal{U}_k$ denotes the upper-triangular pairs in group $k$. The intra-axis loss averages over all six groups:
\begin{equation}
    \mathcal{L}_{\text{Intra}} = \frac{1}{6} \sum_{k=1}^{6} \bigl( \mathcal{L}_{\text{KL}}^{(k)} + \mathcal{L}_{\text{margin}}^{(k)} \bigr),
    \label{eq:loss_intra}
\end{equation}
For the \textit{inter-axis} loss, to provide a supplementary regularization signal for diagonal pairs (\textit{e.g.,} $m_0$ vs.\ $m_8$, which differ along both urgency and comfort), we apply the same pairwise KL-margin loss over all $M{=}9$ personas:
\begin{equation}
    \mathcal{L}_{\text{Inter}} = \mathcal{L}_{\text{KL}}^{(\text{all})} + \mathcal{L}_{\text{margin}}^{(\text{all})},
    \label{eq:loss_inter}
\end{equation}
The final Axis-Decomposed Diversity Loss combines both components:
\begin{equation}
    \mathcal{L}_{\text{AD}} = \mathcal{L}_{\text{Intra}} + w_{\text{Inter}} \cdot \mathcal{L}_{\text{Inter}},
    \label{eq:loss_ad}
\end{equation}
where $w_{\text{Inter}} = 0.2$. The intra-axis term provides clear, unambiguous learning signals by forcing diversity among personas that differ along exactly one axis, while the inter-axis term acts as auxiliary regularization to prevent diagonal mode collapse.

\section{Source Code and PCT Dataset}
\color{black}The code and PCT Dataset are available at \url{https://github.com/VisualAIKHU/PersonaDrive}.\color{black}

\setcounter{algorithm}{0}

\begin{figure*}[t]
\begin{tcolorbox}[
  colback=gray!5, colframe=gray!60, 
  title={\textbf{Prompt 1.} PCT Dataset Generation Pipeline},
  fonttitle=\small\bfseries,
  boxrule=0.6pt, arc=2pt, left=4pt, right=4pt, top=3pt, bottom=3pt
]
\small
\textbf{Input:} Scene set $\mathcal{S}$ from NAVSIM with GT trajectories $\mathrm{GT}_i \in \mathbb{R}^{10 \times 3}$; \;
Persona set $\mathcal{P}$ ($3 \times 3$ urgency-comfort grid) \\[2pt]
\textbf{Output:} Per-scene trajectory and persona description $\{(\boldsymbol{\tau}_k^{(i)},\, d_k^{(i)})\}_{k \in \mathcal{P}}$

\tcblower

\textbf{Stage 1: Candidate Generation and Voting} \\[1pt]
For each scene $s_i$: \\
\quad 1. Extract scene context $c_i$, lane path options $\mathcal{L}_i$, and GT statistics $g_i$. \\
\quad 2. Query GPT-4o-mini with \textbf{Prompt 2} to generate five candidate sets of per-persona parameters; select the best through voting. \\
\quad 3. Synthesize waypoint trajectories $\boldsymbol{\tau}_k^{(i)}$ from parameters and lane geometry. \\
\quad 4. Query GPT-4o-mini with \textbf{Prompt 3} to generate persona descriptions $\{d_k^{(i)}\}$ grounded in the trajectory statistics.

\medskip
\textbf{Stage 2: Rule-Based Validation} \\[1pt]
\quad$\bullet$ Speed ordering, TTC $\geq 1.0$\,s, diversity ADE $\geq 0.3$\,m, trajectory reversal, direction consistency, and text constraints.

\medskip
\textbf{Stage 3: LLM-as-a-Judge} \\[1pt]
\quad$\bullet$ GPT-4o scores each candidate via a trajectory judge (\textbf{Prompt 4}; 9 criteria) and a text judge (\textbf{Prompt 5}; classification + 4 criteria). Labels are blind-shuffled to prevent bias. A sample passes only when the combined score exceeds the threshold.

\medskip
\textbf{Stage 4: Failure-Aware Regeneration} \\[1pt]
\quad$\bullet$ Regenerate failing scenes with failure reasons injected into the prompt (max 3 retries). If validation still fails, the candidate with the highest combined score is selected.
\end{tcolorbox}
\end{figure*}

\begin{figure*}[t]
\begin{tcolorbox}[
  colback=blue!3, colframe=blue!40, 
  title={\textbf{Prompt 2.} Persona-Conditioned Trajectory Prompt},
  fonttitle=\small\bfseries,
  boxrule=0.6pt, arc=2pt, left=4pt, right=4pt, top=3pt, bottom=3pt
]
\small
\textbf{Input:} Scene context $c$, available lane paths $\mathcal{L}$, GT statistics $g$ \\
\textbf{Output:} JSON with per-persona parameters $\{k: \{m_k, f_k, o_k\}\}_{k \in \mathcal{P}}$ \; (structured output, strict mode)

\tcblower

\textbf{System Message} \\[2pt]
\textit{(1) Role:} \\
\quad ``You are a driving behavior simulator embodying 9 radically different human drivers. \\
\quad\; All 9 must produce clearly different trajectories.'' \\[3pt]
\textit{(2) Output parameters} (per persona): \\
\quad Maneuver $m \in \{\texttt{follow},\, \texttt{shift\_left},\, \texttt{shift\_right}\}$, \;
speed factor $f \in \mathbb{R}^{+}$ (vs. GT), \;
lane offset $o \in \mathbb{R}$ (meters) \\[3pt]
\textit{(3) Persona definitions:} \\
\quad Each persona has a character backstory with target ranges for $f$ and $|o|$. \\
\quad Urgency governs speed; comfort governs lateral behavior. \\[3pt]
\textit{(4) Hard constraints:} \\
\quad Monotonic speed ordering ($f_{\text{UH\_CL}} > \cdots > f_{\text{UL\_CH}}$, all unique); \\
\quad Lane change only if adjacent lane available; \;
CH personas: \texttt{follow} with $o{=}0$.

\medskip
\textbf{User Message} \\[2pt]
Serialized scene context (ego state, road geometry, top-10 agents, traffic lights, GT waypoints), \\
available paths from $\mathcal{L}$. \; ``Output ONE JSON object with maneuver, speed\_factor, lane\_offset for each persona.''
\end{tcolorbox}
\end{figure*}

\begin{figure*}[t]
\begin{tcolorbox}[
  colback=orange!3, colframe=orange!40, 
  title={\textbf{Prompt 3.} Persona Description Prompt},
  fonttitle=\small\bfseries,
  boxrule=0.6pt, arc=2pt, left=4pt, right=4pt, top=3pt, bottom=3pt
]
\small
\textbf{Input:} Scene context $c$, trajectory statistics $\{\text{stats}_k\}_{k \in \mathcal{P}}$, situation assignments $h$ \\
\textbf{Output:} JSON with per-persona descriptions $\{k: \{\texttt{user\_persona}\}\}_{k \in \mathcal{P}}$ \; (structured output, strict mode)

\tcblower

\textbf{System Message} \\[2pt]
\textit{(1) Role:} \\
\quad ``You write what a real person would say to an autonomous taxi. \\
\quad\; 9 passengers with different life situations ride through the same scene. \\
\quad\; Natural human speech, not robotic commands.'' \\[3pt]
\textit{(2) Persona-to-situation mapping} (one situation per persona from a curated pool): \\[2pt]
\quad\begin{tabular}[t]{@{}l@{\;\;}l@{\qquad}l@{\;\;}l@{}}
  UH\_CL: & desperate crisis         & UH\_CM: & professional deadline \\
  UH\_CH: & urgent + fragile cargo   & UM\_CL: & casual positional preference \\
  UM\_CM: & mundane errand            & UM\_CH: & fragile passenger \\
  UL\_CL: & slow sightseeing         & UL\_CM: & medical/physical reason \\
  UL\_CH: & genuinely terrified       &         & \\
\end{tabular} \\[4pt]
\textit{(3) Text rules:} \\
\quad 2 to 4 sentences, $\leq 350$ characters, natural spoken English; concrete situation, scene-reactive.

\medskip
\textbf{User Message} \\[2pt]
Compact scene summary, per-persona trajectory statistics, and hash-based situation assignments. \\
``Match trajectory stats (fast $\to$ urgent, slow $\to$ vulnerable). Output ONE JSON object only.''
\end{tcolorbox}
\end{figure*}

\begin{figure*}[t]
\begin{tcolorbox}[
  colback=teal!3, colframe=teal!40, 
  title={\textbf{Prompt 4.} Trajectory Judge Prompt (Stage 3)},
  fonttitle=\small\bfseries,
  boxrule=0.6pt, arc=2pt, left=4pt, right=4pt, top=3pt, bottom=3pt
]
\small
\textbf{Input:} Scene context $c$, nine blind-shuffled trajectories $\{\boldsymbol{\tau}_A, \ldots, \boldsymbol{\tau}_I\}$ \\
\textbf{Output:} Per-trajectory scores (structured JSON, strict mode)

\tcblower

\textbf{System Message} \\[2pt]
\textit{(1) Role:}
``Expert driving trajectory evaluator. Receives a scene context and nine anonymized trajectories (A--I) from the same scene.'' \\[3pt]
\textit{(2) Evaluation criteria} (integer score 1--5 per trajectory): \\[1pt]
\quad \texttt{speed\_level} (vs.\ GT speed), \;
\texttt{aggressiveness}, \;
\texttt{smoothness}, \;
\texttt{safety}, \\
\quad \texttt{scene\_fit}, \;
\texttt{naturalness}, \;
\texttt{traffic\_rule\_compliance}, \\
\quad \texttt{maneuver\_plausibility}, \;
\texttt{temporal\_consistency}

\medskip
\textbf{User Message} \\[2pt]
Scene context (ego speed, road geometry, lead vehicles, traffic lights, GT reference speed) and per-trajectory waypoint coordinates. Labels are blind-shuffled to prevent bias.
\end{tcolorbox}
\end{figure*}

\begin{figure*}[t]
\begin{tcolorbox}[
  colback=red!3, colframe=red!40, 
  title={\textbf{Prompt 5.} Text Judge Prompt (Stage 3)},
  fonttitle=\small\bfseries,
  boxrule=0.6pt, arc=2pt, left=4pt, right=4pt, top=3pt, bottom=3pt
]
\small
\textbf{Input:} Scene context $c$, nine blind-shuffled persona descriptions $\{d_A, \ldots, d_I\}$ \\
\textbf{Output:} Per-text classification and quality scores (structured JSON, strict mode)

\tcblower

\textbf{System Message} \\[2pt]
\textit{(1) Role:}
``Expert evaluator of passenger request texts for autonomous taxis. Receives nine anonymized descriptions (A--I) from the same scene.'' \\[3pt]
\textit{(2) Classification} (per text): \\
\quad Urgency: \texttt{HIGH} / \texttt{MEDIUM} / \texttt{LOW}; \;
Comfort: \texttt{AGGRESSIVE} / \texttt{MODERATE} / \texttt{SMOOTH} \\[3pt]
\textit{(3) Quality criteria} (integer score 1--5 per text): \\[1pt]
\quad \texttt{naturalness}, \;
\texttt{plausibility}, \;
\texttt{scene\_reactivity}, \;
\texttt{specificity}

\medskip
\textbf{User Message} \\[2pt]
Scene context and nine persona description texts. Labels are blind-shuffled to prevent bias.
\end{tcolorbox}
\end{figure*}

\end{document}